\documentclass[preprint,5p,times]{elsarticle}

\usepackage{amsmath,amssymb,amsfonts}
\usepackage{algorithmic}
\usepackage{textcomp}
\usepackage{soul}
\usepackage{float}
\usepackage{caption}
\usepackage{subcaption}
\usepackage{bm}
\usepackage{comment}
\usepackage{algorithm}
\usepackage{hyperref}

\usepackage{booktabs}
\usepackage{multirow}
\usepackage{graphicx}
\usepackage{array}
\usepackage{arydshln}
\usepackage[htt]{hyphenat}

\journal{Computers in Biology and Medicine}

\begin{document}

\begin{frontmatter}



\title{Logic Gate Networks and Lookup Table Networks as Lightweight Hardware Classifiers for Inter-patient ECG Arrhythmia Classification}

\author[imec,vub]{Wout Mommen\corref{cor1}}
\ead{wout.mommen@imec.be}
\cortext[cor1]{Corresponding author.}
\author[imec]{Lars Keuninckx} 
\ead{lars.keuninckx@imec.be}
\author[imec]{Siddharth Patil} 
\author[imec_nl]{Paul Detterer} 
\ead{paul.detterer@imec.nl}
\author[imec]{Achiel Colpaert} 
\ead{achiel.colpaert@imec.be}
\author[imec,vub]{Piet Wambacq} 
\ead{piet.wambacq@imec.be}

\affiliation[imec]{organization={Interuniversity Microelectronics Centre (IMEC)},
            addressline={Kapeldreef 75}, 
            city={Leuven},
            postcode={3001}, 
            country={Belgium}}
            
\affiliation[vub]{organization={Vrije Universiteit Brussel},
            addressline={Pleinlaan 2}, 
            city={Elsene},
            postcode={1050}, 
            country={Belgium}}
\affiliation[imec_nl]{organization={imec Holst Centre},
            addressline={High Tech Campus 31}, 
            city={Eindhoven},
            postcode={5656 AE}, 
            country={the Netherlands}}

\begin{abstract}
Deep Differentiable Logic Gate Networks (LGNs) and Lookup Table Networks (LUTNs) offer a promising approach for very low power inference due to their use of simple binary logic operations instead of arithmetic. In this work, we generalize the logic gates of LGNs to more than two input pins, naturally arriving at networks consisting of $N$-input LUTs. To obtain a differentiable expression for training the $N$-LUT entries, we adopt the Boolean equation of a $2^N$:1 multiplexer (MUX) and optimize its input parameters during training. We investigate the applicability of LGNs and LUTNs to inter-patient ECG arrhythmia classification using the MIT-BIH data set. The proposed models achieve up to 94.41\% accuracy and a $j\kappa$ index of 0.683 on a four-class task, showing a competitive performance compared to existing CNN-, SVM- and SNN-based methods. Our LGNs and LUTNs only require an estimated 2.89k to 6.17k FLOPs, including preprocessing and readout, which is three to six orders of magnitude less than state-of-the-art methods. We verified our design, which consists of the preprocessing pipeline and a 6-LUTN classifier, by implementing it on a Xilinx Zynq-7000 ZedBoard. The complete system consumes a dynamic energy of 8.25 µJ/inference, of which only 0.46 nJ is utilized by the LUTN classifier. These results show that both LGNs and LUTNs can be employed as lightweight hardware-based classifiers for inter-patient ECG arrhythmia classification.
\end{abstract}




\end{frontmatter}



\section{Introduction}
\label{sec:introduction}
The World Health Organization (WHO) states that cardiovascular diseases (CVDs) are the leading cause of death globally, accounting for  nearly 17.9 million deaths each year \cite{world_health_organization_who_cardiovascular_2025}. Hence, measures need to be taken for early detection or monitoring of such CVDs. Heart implants or even wearables could obtain the necessary information, provided that the \linebreak[4] battery-operated hardware uses very low power. Often, the MIT-BIH arrhythmia data set \cite{moody_impact_2001} is used as a first step to benchmark different technologies or algorithms for ECG arrhythmia classification, irrespective of the power usage \cite{mondejar-guerra_heartbeat_2019,chazal_automatic_2004,li_inter-patient_2022,villa_are_2019}. However, if these algorithms are envisioned to run on small embedded devices such as pacemakers, where having to replace the battery is undesirable, care should be taken regarding the power and energy consumption of these solutions.\\

Quantization and pruning of artificial neural networks have been used in the past to increase power and energy efficiency of these algorithms \cite{hawks_ps_2021}. A next approach is to use a Spiking Neural Network (SNN) to realize the arrhythmia classification \cite{mao_ultra-energy-efficient_2022,yan_energy_2021}. To further reduce energy consumption, researchers have explored non-conventional computing systems. For example, the use of Monostable Multivibrator (MMV) networks has been shown to improve power efficiency and is straightforward to implement in hardware \cite{keuninckx_training_2025}. These MMV networks use simple digital timers as their basic element. Building on this trend, recent studies have demonstrated that the network elements can be reduced even further, enabling trainable networks of digital logic gates to tackle a wide range of tasks. These Deep Differentiable Logic Gate Networks (LGNs) \cite{petersen_deep_2022} are constructed from a 2-input logic gate network configuration with randomly chosen fixed connections. Given a certain data set, the optimal gate types are learned utilizing traditional backpropagation techniques. Similarly, networks built from simple Lookup Tables (LUTs) have been proposed in literature under different names: LUTNet \cite{wang_lutnet_2019,wang_lutnet_2020}, LogicNets \cite{umuroglu_logicnets_2020}, NullaNet \cite{nazemi_energy-efficient_2019}, PolyLUT \cite{andronic_polylut_2023}, NeuraLUT \cite{andronic_neuralut_2024}, Differentiable Weightless Neural Networks (DWNs) \cite{bacellar_differentiable_2024} and Walsh-Assisted Relaxation for Probabilistic Look-Up Tables (WARP-LUTs) \cite{gerlach_warp-luts_2025}. For clarity, in this work, we will use the term Lookup Table Network (LUTN) since this best represents its essence. Although a LUT element consists of multiple logic gates, it offers two advantages over LGNs. Firstly, a LUT is a reconfigurable element that enables a single LUTN architecture to execute different functionalities. Secondly, a LUT can have more than two input ports, allowing for more complex neuron functions. To create a LUTN for a specific application, two methods have been presented. First, a regular neural network can be converted into a LUTN using a multitude of techniques \cite{wang_lutnet_2019,wang_lutnet_2020,umuroglu_logicnets_2020,nazemi_energy-efficient_2019,andronic_polylut_2023,andronic_neuralut_2024}. Alternatively, a LUTN can be trained directly \cite{bacellar_differentiable_2024,gerlach_warp-luts_2025}. The first solution allows for conventional neural network training, but the final accuracy is lower and the number of LUTs is higher compared to directly trainable LUT networks. Questions remain open on whether a LUTN can be used for efficient arrhythmia classification, what potential data preprocessing would be required, and how to efficiently train these LUT networks for this type of application.\\

In this work, to address the aforementioned open questions, we propose an ECG classification system using LUTNs by implementing a novel training method and preprocessing method. The contributions of this work are:
\begin{enumerate}
    \item We present a systematic evaluation of Deep Differentiable Logic Gate Networks (LGNs) and Lookup Table Networks (LUTNs) for ECG arrhythmia classification using the inter-patient training and testing paradigm. This evaluation is compliant with the guidelines of the Association for the Advancement of Medical Instrumentation (AAMI). Although the LGNs have been explored for ECG classification in prior work, this study focuses on a mixed-patient setting. In contrast, this work analyses the LGNs and LUTNs under the more challenging inter-patient paradigm, which better represents real-world scenarios.
    
    \item The models that are studied obtain a competitive performance for the four-class MIT-BIH classification task, \linebreak[4] achieving competitive results compared to CNN-, SVM-, SNN-based models using comparable evaluation methods, while requiring only an estimated 2.89k to 6.17k operations, including preprocessing and readout. 

    \item We analyse the behaviour and limitations of LGN-based and LUT-based models for ECG classification, including LUT input size, network depth, and rate coding, identifying the practical regimes in which these models operate effectively.

    \item We show that logic-based ECG classifiers can be implemented on FPGA hardware in an efficient and straightforward manner. Rather than targeting a specific deployment platform, the FPGA implementation serves as a verification for our preprocessing and classification methodologies. The estimated power consumption, resource utilization, and latency are reported. These results provide an early indication that our proposed approach has potential for ultra-low-complexity edge deployment.
\end{enumerate}

\section{Related work}
 Spiking Neural Networks (SNNs) \cite{maass_networks_1997} are often adopted as energy-efficient AI models for a wide range of applications, including ECG arrhythmia classification. Mao et al. \cite{mao_ultra-energy-efficient_2022} benchmark the MIT-BIH data set on an SNN, achieving an accuracy of 93.67\% using 0.3~µJ per inference on a chip with a leakage power of 1.14 µW. Yan et al. \cite{yan_energy_2021} trained a two-stage CNN on the MIT-BIH data set. The first classifier acts as a wake-up system to determine if there is a normal heartbeat or an arrhythmia and the second classifier determines the type of arrhythmia. Both CNNs were converted to SNNs using rate coding, achieving an accuracy of 90.00\% and a j$\kappa$ index of 0.609. The j$\kappa$ index is a figure of merit commonly used for the MIT-BIH data set, where higher values indicate a better classification performance. The estimated power to run these SNNs back to back is 77 mW.\\

Deep Differentiable Logic Gate Networks (LGNs) were first introduced by Petersen et al. \cite{petersen_deep_2022}. An LGN is a network of randomly connected 1-input and 2-input logic gates. The gate layers are connected to each other using randomly chosen connections that remain fixed during training. The gate type, on the other hand, is learned using backpropagation by creating a linear combination of the operations $f_i$ of all 16 gate types and learning the corresponding coefficients $w_i$. One can apply the softmax function on the coefficients, essentially learning a discrete probability distribution $p_i$ over all possible gate types. The output of the combination of gate types (i.e. an LGN ``neuron'') is the activation and is given by \cite{petersen_deep_2022}

\begin{equation} \label{eq:activation_LGN}
a=\sum_{i=0}^{15} \boldsymbol{p}_i \cdot f_i\left(x_0, x_1\right)=\sum_{i=0}^{15} \frac{e^{\boldsymbol{w}_i}}{\sum_j e^{\boldsymbol{w}_j}} \cdot f_i\left(x_0, x_1\right).
\end{equation}

The real-valued logic operations $f_i$ used in this formula can be found in Table \ref{tab:LGN}. Since Eq. \eqref{eq:activation_LGN} is differentiable, backpropagation is utilized to learn the most optimal gate type for each gate in the network. To finalize the training procedure, the gate type with the highest probability is chosen for each gate-type superposition. This final network is then employed for inference. After the final layer of the network, population counters are utilized to determine the predicted class label. The population counters are constructed as follows: First, the number of gates $G$ in the final layer, needs to be dividable by the number of classes $C$. Next, the output values of gate 1 until gate $G/C$ are summed, which becomes the population count of the first class. For the population count of the second class, the gate outputs from gate $G/C+1$ to $2G/C$ are summed. The population counts of the other classes are calculated in a similar manner, i.e. for class $i$, the gate output values of gate $(i-1)\cdot G/C + 1$ until gate $i\cdot G/C$, with $i=1,\ldots,C$, are summed. The highest value of these population count values or group sums determines the predicted class label. During training, these population count values are converted to probabilities using a softmax function with a temperature scaling factor $T$, which is a hyperparameter of the model. Besides feedforward LGNs, it is also possible to create recurrent \cite{buhrer_recurrent_2025} and convolutional architectures \cite{petersen_convolutional_2024}. All of these architectures exhibit very low power consumption at a high throughput, as they are essentially networks of 2-input logic gates. It is also possible to train the connections of these networks \cite{mommen_method_2025}, resulting in a much smaller network with a similar performance. Apart from the classical LGNs, Gumbel LGNs \cite{yousefi_mind_2025} have been proposed, which use Gumbel noise to decrease training time, increase the number of \textit{used} neurons, and decrease the discretization gap. Here, the discretization gap is a small drop in accuracy that occurs when we choose the gate type with the highest probability value for each gate-type superposition in Eq. \eqref{eq:activation_LGN}.\\

\begin{table}[t]
    \centering
    \caption{The index of summation $i$, corresponding binary operation, truth table and real-valued logic operation $f_i$ utilized in Eq. \eqref{eq:activation_LGN} \cite{petersen_deep_2022}.}
    \label{tab:LGN}
    \resizebox{\columnwidth}{!}{%
    \begin{tabular}{ccccccl}
        \toprule 
        $i$    & Operation                         & 00 & 01 & 10 & 11 & $f_i$                   \\ \midrule
        0      & False                             & 0  & 0  & 0  & 0  & 0                       \\
        1      & $x_0$ $\wedge$ $x_1$              & 0  & 0  & 0  & 1  & $x_0x_1$                \\
        2      & $\neg$($x_0$ $\Rightarrow$ $x_1$) & 0  & 0  & 1  & 0  & $x_0-x_0x_1$            \\
        3      & $x_0$                             & 0  & 0  & 1  & 1  & $x_0$                   \\
        4      & $\neg$($x_0$ $\Leftarrow$ $x_1$)  & 0  & 1  & 0  & 0  & $x_1-x_0x_1$            \\
        5      & $x_1$                             & 0  & 1  & 0  & 1  & $x_1$                   \\
        6      & $x_0$ $\oplus$ $x_1$              & 0  & 1  & 1  & 0  & $x_0+x_1-2 x_0 x_1$     \\
        7      & $x_0$ $\vee$ $x_1$                & 0  & 1  & 1  & 1  & $x_0+x_1-x_0 x_1$       \\ \hdashline
        8      & $\neg$($x_0$ $\vee$ $x_1$)        & 1  & 0  & 0  & 0  & $1-(x_0+x_1-x_0 x_1)$   \\
        9      & $\neg$($x_0$ $\oplus$ $x_1$)      & 1  & 0  & 0  & 1  & $1-(x_0+x_1-2 x_0 x_1)$ \\
        10     & $\neg$ $x_1$                      & 1  & 0  & 1  & 0  & $1-x_1$                 \\
        11     & $x_0$ $\Leftarrow x_1$            & 1  & 0  & 1  & 1  & $1-x_1+x_0 x_1$         \\
        12     & $\neg$ $x_0$                      & 1  & 1  & 0  & 0  & $1-x_0$                 \\
        13     & $x_0$ $\Rightarrow$ $x_1$         & 1  & 1  & 0  & 1  & $1-x_0+x_0 x_1$         \\
        14     & $\neg$ ($x_0$ $\wedge$ $x_1$)     & 1  & 1  & 1  & 0  & $1-x_0 x_1$             \\
        15     & True                              & 1  & 1  & 1  & 1  & 1                       \\ \bottomrule
        \end{tabular}}
\end{table}

Lookup Table Networks (LUTNs) \cite{bacellar_differentiable_2024} are similar to LGNs, but instead of using logic gates as ``neurons'', Lookup Tables (LUTs) are employed. These LUTNs can be seen as a generalization of LGNs, in the sense that a LUT transforms an N-bit input value into one binary output value, while a logic gate transforms a 2-bit input value into one binary output value. Hence, a 2-LUTN is equivalent to an LGN. Due to the increased complexity of the ``neurons'', an $N$-input LUT can learn exponentially more functions compared to a logic gate. Namely, the number of possible functions scales as $2^{2^N}$, as can be seen from Table \ref{tab:LUT_func}. The most basic building block of most FPGAs is the 6-LUT. Thus, having the ability to directly learn a network of 6-LUTs on a specific data set can lead to very efficient implementations of AI algorithms on FPGAs. Additionally, due to the high complexity of these 6-LUTs compared to logic gates, it is expected that we will need fewer neurons (i.e. LUT primitives) and hence less global interconnect for chip implementations. Consequently, chips could be smaller and routing easier when using 6-LUTs compared to logic gates. Along with reconfigurability, this is the primary motivation for switching from LGNs to LUTNs.\\

Some research has already been conducted on LUTNs. Wang et al. \cite{wang_lutnet_2019,wang_lutnet_2020} take a pruned XNOR-Net \cite{rastegari_xnor-net_2016} and convert it to a network of LUTs, called LUTNet. This network still includes activation functions and accumulation operations. Umuroglu et al. introduce LogicNets \cite{umuroglu_logicnets_2020}, which are sparse quantized neural networks converted into networks of LUTs. They manage to map an artificial neuron with limited and quantized inputs (X) and outputs (Y) to an X:Y logical LUT, which can be implemented on FPGAs using their underlying hardware primitives, such as 6:1 or 5:2 physical LUTs. The transformation from neuron to logical LUT eliminates the need for multiply and accumulate operations, making the computation more efficient. The final model only consists of physical LUTs and the connectivity between them. Nazemi et al. introduce NullaNet \cite{nazemi_energy-efficient_2019}, and also convert neurons to logical LUTs. Conversely, these logical LUTs are converted to sum-of-products terms and are next converted to logic gates using a logic synthesis tool, which optimizes for area, delay, and power consumption. Andronic et al. propose PolyLUT \cite{andronic_polylut_2023}, which is similar to LogicNets, since it also maps neurons to logical and physical LUTs. However, their neuron contains multivariate polynomials instead of a linear transformation. In that way, fewer physical LUTs are needed to match a similar accuracy. Following this, Andronic et al. present NeuraLUT \cite{andronic_neuralut_2024}, that builds further on PolyLUT, modelling entire sub-networks in the same size logical LUT. Previous methods train conventional neural networks and convert them to networks of LUTs; yet, research has also been conducted on directly constructing networks of LUTs. Bacellar et al. \cite{bacellar_differentiable_2024} train both the connections and LUT entries in a network of LUTs they call a Differentiable Weightless Neural Network (DWN). They utilize an extended finite difference method to train the LUT. Gerlach et al. \cite{gerlach_warp-luts_2025} continue on the works of the original LGNs and Gumbel LGNs towards an algorithm called Walsh-Assisted Relaxation for Probabilistic Look-Up Tables (WARP-LUTs). These are networks of 2-input LUTs that are trained with backpropagation, but require fewer parameters and training time compared to LGNs and Gumbel LGNs. As of the time of writing, networks consisting of LUTs with more than two input pins have not been tried out using this method.\\

\begin{table}[H]
    \centering
    \caption{The primitive and number of binary operations it can express. An $N$-LUT stands for a lookup table with $N$ input pins.}
    \label{tab:LUT_func}
        \begin{tabular}{ll}
        \toprule
        Primitive & No. binary functions \\
        \midrule
        2-LUT & $16$ \\
        4-LUT & $65\ 536$ \\
        6-LUT & $1.845 \cdot 10^{19}$ \\
        $N$-LUT & $2^{2^N}$ \\ \bottomrule
        \end{tabular}
\end{table}

\section{Methodology}

    In this section, we will first describe the data set and which part of the data is used for training and testing. Following this, we explain the novel preprocessing method for extracting the features that are the inputs to the models. Subsequently, we describe the training method for LGNs and how rate coding is applied to improve performance. Next, the training method of the LUTNs is delineated. Finally, we describe the evaluation metrics and experimental protocol that are adopted in this work.

    \subsection{Preprocessing} \label{sec:preprocessing}
    \subsubsection{MIT-BIH data set}
    The data set contains 48 two-channel ECG recordings, each about 30 minutes long, originating from 47 patients. This results in approximately 100000 heartbeats, containing both normal heartbeats and various arrhythmias \cite{moody_impact_2001}. The signal was recorded at a sample rate of 360 Hz  with a resolution of 11 bits, and independently labeled by two or more cardiologists. We only utilized one channel and omit certain parts of the data as explained in the next subsection.\\
        \subsubsection{Train-test split}
        The data set was split into a train set (DS1) and test set (DS2) according to the inter-patient scheme designed by \cite{chazal_automatic_2004}:
        \begin{itemize}
            \item DS1: 101, 106, 108, 109, 112, 114, 115, 116, 118, 119, 122, 124, 201, 203, 205, 207, 208, 209, 215, 220, 223, 230
            \item DS2: 100, 103, 105, 111, 113, 117, 121, 123, 200, 202, 210, 212, 213, 214, 219, 221, 222, 228, 231, 232, 233, 234
        \end{itemize}
        Here, each number designates a patient recording, where one channel of the two-channel signal was used, following the protocol employed in \cite{mondejar-guerra_heartbeat_2019,li_inter-patient_2022}. As proposed by the Association for the Advancement of Medical Instrumentation (AAMI), patient records containing paced beats (102, 104, 107, 217) were removed \cite{aami}. Additionally, beats were grouped together in new classes according to the AAMI guidelines \cite{aami}:
        
        \begin{align*}
            \text{N, L, R} &\rightarrow \text{N} & (89\,474=45\,626+43\,848)\\
            \text{e, j, A, a, J, S} &\rightarrow \text{S} & (6\,986 = 3\,778 + 3\,208)\\
            \text{V, E} &\rightarrow \text{V} & (3\,018 = 975 + 2\,043)\\
            \text{F} &\rightarrow \text{F} & (801 = 413 + 388)\\
            \text{Q,/,f} &\rightarrow \text{Q} & (15 = 8 + 7).
        \end{align*}
        Following these guidelines as in \cite{mar_optimization_2011,mondejar-guerra_heartbeat_2019,zhang_heartbeat_2014}, the Q class is ignored, since there are only 15 examples in the data set (8 train examples and 7 test examples). This results in a four-class classification problem: normal heartbeats (N), supraventricular ectopic beats (S or SVEB), ventricular ectopic beats (V or VEB), and beats resulting from the fusion of VEBs and normal beats (F) \cite{chazal_automatic_2004}.
        
        \subsubsection{Feature extraction} \label{sec:feature_extraction}
        The goal of the feature extraction step is to convert the full-precision raw ECG data into meaningful (binary) features from which the classifier can distinguish the four different classes more easily. Since both LGNs and LUTNs work with binary input values, a binary feature vector of length 138 was constructed containing the following parameters:
        \begin{align} \label{eq:feature_vector}
            &[RR_1, RR_2, RR_3, RR_4, \Delta RR_p, \Delta RR_m, \\ \nonumber 
            &\quad RR_{locCV}, RR_{ratio}, t_b,M_1,M_2,M_4,cf_1,cf_2,\delta]. 
        \end{align}
        Each of these features is explained in the following paragraphs.\\
        
\begin{figure}[t]
            \centering
             \includegraphics[width=0.5\textwidth,keepaspectratio]{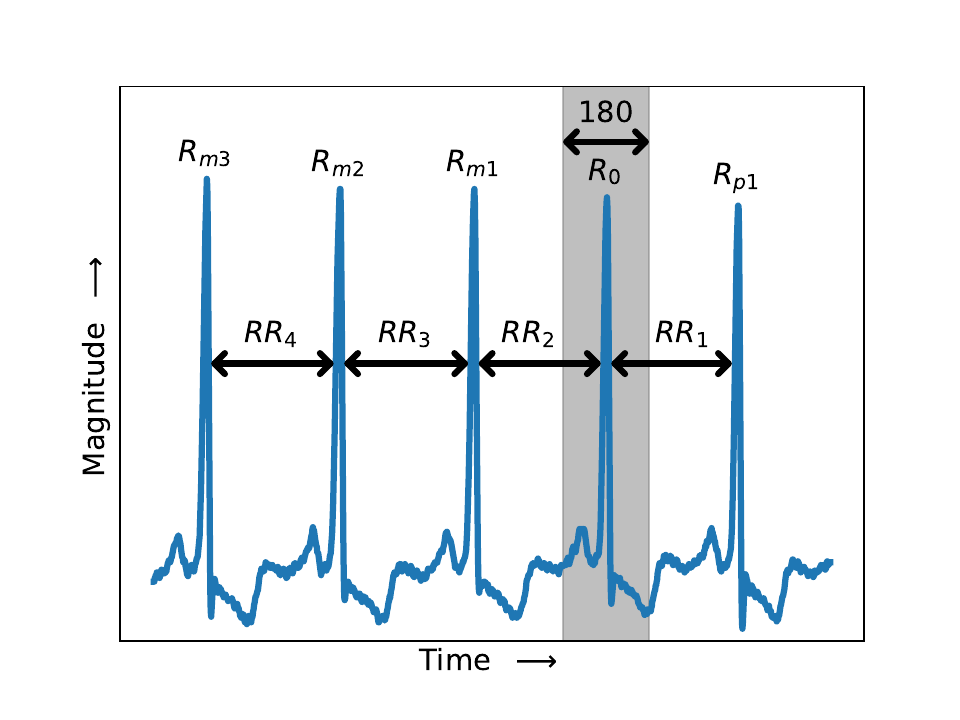}
             \caption{Excerpt of the ECG data highlighting the current R-peak position $R_0$, three previous R-peak positions $R_{m1}$, $R_{m2}$, $R_{m3}$, and the next R-peak position $R_{p1}$. Using these peak positions, all four RR-interval features were derived. Around each peak, we define a gray window of 180 samples that was utilized to determine the morphological features $M_1$ to $M_4$ and the crest factor $cf_1$.}
             \label{fig:preprocessing}
        \end{figure}

        First, for each peak position $R_0$, the next peak position $R_{p1}$ and three previous peak positions $R_{m1}$, $R_{m2}$ and $R_{m3}$ were extracted as seen in Fig. \ref{fig:preprocessing}. The peak positions originate from the MIT-BIH data set itself, hence no peak detection algorithm was utilized. Using these peak positions the RR-intervals \cite{mondejar-guerra_heartbeat_2019,mao_ultra-energy-efficient_2022} $RR_1$, $RR_2$, $RR_3$ and $RR_4$ were determined. Each of the intervals was encoded into an 8-bit binary number. 
        The features $\Delta RR_p$ and $\Delta RR_m$ designate the RR-interval changes \cite{mao_ultra-energy-efficient_2022} and are simply sign bits:
        \begin{align*}
            \Delta RR_p &= \begin{cases}1 & \text { if } RR_1 > RR_2 \\ 0 & \text { otherwise }\end{cases} \\
            \Delta RR_m &= \begin{cases}1 & \text { if } RR_2 > RR_3 \\ 0 & \text { otherwise. }\end{cases} \\
        \end{align*}

        Inspired by \cite{mondejar-guerra_heartbeat_2019}, the local RR interval was determined, specifically $RR_2$ of $n=500$ previous examples was calculated. Following this, the mean $m$ and standard deviation $s$ of this list of numbers were calculated to obtain the coefficient of variation $s/m$. To save power, this value was updated for each heartbeat, rather than calculating it fully anew at every beat. The coefficient of variation $s/m$ was converted to binary numbers by using thresholds of 0.5 and 0.1 to obtain a two-bit feature $RR_{locCV}$ for each heartbeat. Additionally, the RR-ratio $RR_{ratio} = RR_1/m$ was calculated and converted to two binary values by comparing its value to thresholds 0.25 and 0.5.\\

        Another feature to be extracted was the tachycardia bit $t_b$, to assess if the currently selected heartbeat occurs at a rate faster than 100 bpm, which was determined using the local RR-interval.\\
        
        Hereafter, morphological features were extracted in a similar manner as \cite{mondejar-guerra_heartbeat_2019}: 
        \begin{itemize}
            \item $M_1 =\left|\text{ecg}[R_0]-\text{min}(\text{beat}[0:40])\right|/\text{norm}$
            \item $M_2 =\left|\text{ecg}[R_0]-\text{min}(\text{beat}[65:85])\right|/\text{norm}$
            \item $M_3 =\left|\text{ecg}[R_0]-\text{min}(\text{beat}[95:105])\right|/\text{norm}$
            \item $M_4 =\left|\text{ecg}[R_0]-\text{min}(\text{beat}[150:180])\right|/\text{norm}.$
        \end{itemize}
        Here, ecg is the array containing the values of a patient's recording, beat is an array containing 180 values around the center peak $R_0$ representing one heartbeat, and $\text{norm} = \text{max}(\text{beat}) - \text{min}(\text{beat})$. For the final feature vector, $M_3$ was excluded, since this increased the accuracy. All these features were encoded into 3 bits.\\

        Additionally, the crest factor was determined for better distinguishing normal beats from ventricular ectopic beats. The crest factor is defined as the peak amplitude of the complete heartbeat window divided by its RMS value. The crest factor $cf_1$ was determined for a window of 180 samples, and $cf_2$ for a window of 400 samples centered at $R_0$. Each crest factor was encoded into 8 bits.\\

        Finally, delta encoding was applied as in \cite{mao_ultra-energy-efficient_2022}, obtaining 74 binary numbers represented by $\delta$ in the feature vector of Eq.~\eqref{eq:feature_vector}.
 
    \subsection{Deep Differentiable Logic Gate Networks}
\begin{figure}[t]
        \centering
         \includegraphics[width=0.5\textwidth,keepaspectratio]{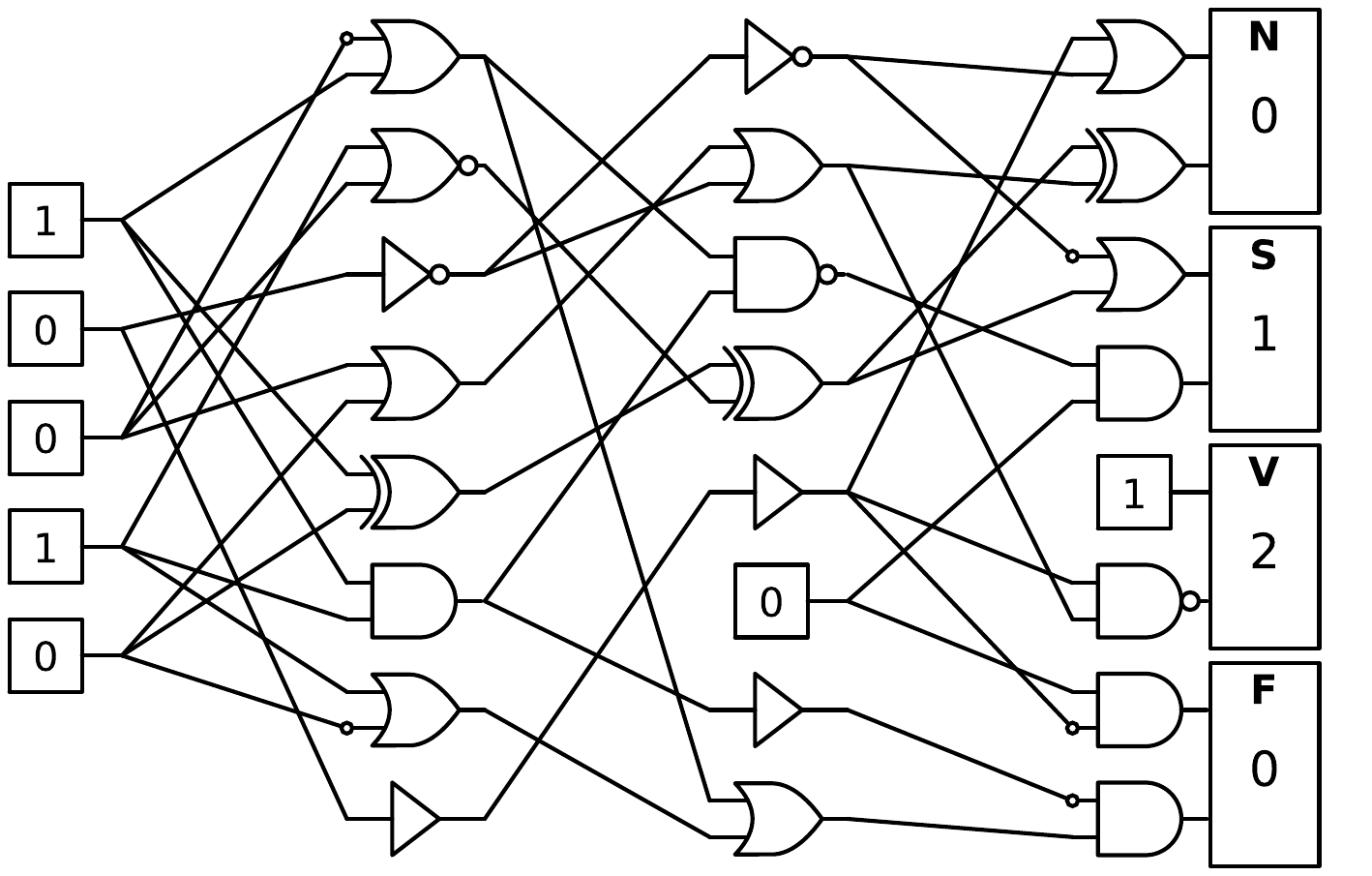}
         \caption{A visualization of a 3-layer LGN that predicts four classes: N, S, V and F. The input values are binary numbers, and are propagated through the network. The final layer groups the gates, and counts the number of 1's that appear at the gate-outputs for each group. The group sum (i.e. popcount) that has accumulated the highest value is the winner and assigns the corresponding class label to the example. In reality, we employ LGNs with 8000 gates per layer instead of just eight as displayed in the figure, meaning that the maximum popcount value is equal to 2000 instead of two.}
         \label{fig:LGN}
\end{figure}

    The Deep Differentiable Logic Gate Networks (LGNs) were trained in the same manner as described in \cite{petersen_deep_2022} with fixed connections and a learning rate of 0.01. All networks had 8000 gates per layer and were trained for 200 epochs using a batch size of 100. A visualization of a 3-layer LGN that predicts the four classes N, S, V and F is shown in Fig. \ref{fig:LGN}. This illustration shows the binary inputs, feedforward gate network, and population counts for the different classes at the output. In the figure, we only have two final-layer gates per popcount, therefore the maximum popcount value is equal to two. However, in this work, we employ 8000 gates per layer, such that the maximum popcount value equals 2000. While training the LGN, the gate types are learned using the differentiable expressions from Table \ref{tab:LGN}. At inference, the inputs to the gates are binary values. Yet, during training, $x_j \in [0,1]$ with $j=1,2$ in these expressions are interpreted as probabilities or rates. More formally, if $X_0,X_1\in\{0,1\}$ are binary random variables, then the probabilistic interpretation states that $P(X_0=1)=x_0$ and $P(X_1=1)=x_1$, where $x_0$ and $x_1$ are rates. To obtain the expressions in Table~\ref{tab:LGN}, the rates, which correspond to bit streams, need to be independent. For example, the output rate of an XOR gate is given by
    \begin{align*}
        P(X_0 \oplus X_1=1) &= P(X_0=1,X_1=0) + P(X_0=0,X_1=1) \\ 
        &= P(X_0=1)P(X_1=0) + P(X_0=0)P(X_1=1) \\ 
        &= x_0(1-x_1) + (1-x_0)x_1 \\ 
        &= x_0+x_1-2x_0x_1,
    \end{align*}
    matching the expression in Table \ref{tab:LGN} and which assumes independent probabilities or bit streams. The other Boolean operations in Table \ref{tab:LGN} follow the same reasoning. Crucially, during training, there is no need to convert the inputs into bit streams; we simply propagate full-precision values between 0 and 1 that represent infinitely long bit streams as part of rate coding. This method not only makes training very fast compared to employing bit streams, but it also ensures that the model achieves the best possible performance when using rate coding. Of course, during inference, one has to use finite-length bit streams, usually resulting in a slightly lower performance depending on the length of the bit stream. In the limit of infinitely long bit streams, the discrete logic gate operations will converge to $f_i$. The full-precision feature values of Eq. \eqref{eq:feature_vector} were used as inputs for the networks using rate coding. Furthermore, an additional RR ratio, $RR_2/m$ is added, since this increased the performance, resulting in 89 input features. For networks with non-rate-coded inputs, the conventional way of training these networks was employed by utilizing the binary input vector Eq. \eqref{eq:feature_vector}.
    
    \subsection{Lookup Table Networks}
    \subsubsection{Proposed training method}
\begin{figure}[t]
        \centering
         \includegraphics[width=0.5\textwidth,keepaspectratio]{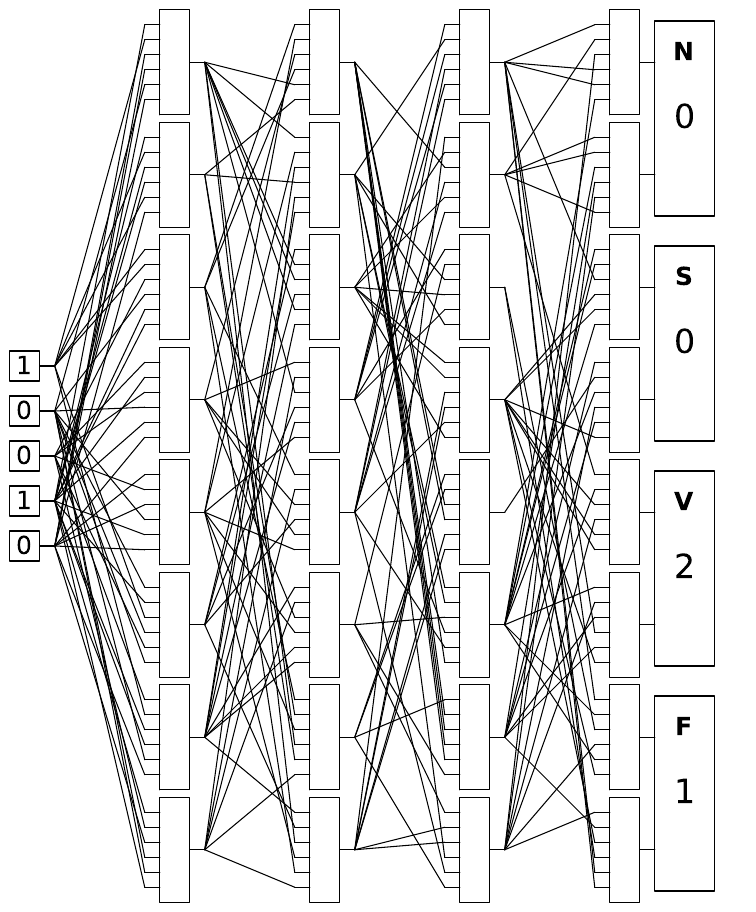}
         \caption{An illustration of a 4-layer Lookup Table Network (LUTN) that consists out of 6-input LUTs. The binary input values are propagated through the LUT layers, where each hidden layer performs a learned binary transformation on its received binary input values. The LUTs in the final layer are grouped. Each group corresponds with a popcount that counts the number of 1's. The popcount with the highest value gives the predicted class label. In this illustration, only eight 6-LUTs per layer are displayed, such that the maximum popcount value is equal to two. For the MIT-BIH data set, we utilize 2000 6-LUTs per layer, such that the maximum popcount value is equal to 500.}
         \label{fig:lutn}
    \end{figure}

    A Lookup Table Network (LUTN) is a network that consists of $N$-input LUTs, arranged in layers. Each LUT receives $N$ binary input values and produces one binary output value. Fig. \ref{fig:lutn} displays an example of a 4-layer LUTN consisting of 6-LUTs. The binary input values are propagated through the network layers, and similarly to the LGNs, the group sum with the highest value gives the predicted class label. In essence, a LUTN is the same as an LGN, but now the 2-input logic gates are replaced with more general $N$-input LUTs.\\
    
    The method for training the LUTNs is based on the fact that a LUT in an FPGA is implemented as a MUX. As an example, the Boolean equation of an $8$:$1$ MUX is given by the following equation:
    
    \begin{equation} \label{eq:3-LUT}
        L_{out} = W_0 \overline{L_0L_1L_2} + W_1 \overline{L_0L_1}L_2 + \ldots + W_7 L_0L_1L_2.
    \end{equation}
    
\begin{figure}[t]
     \centering
     \includegraphics[width=0.5\textwidth]{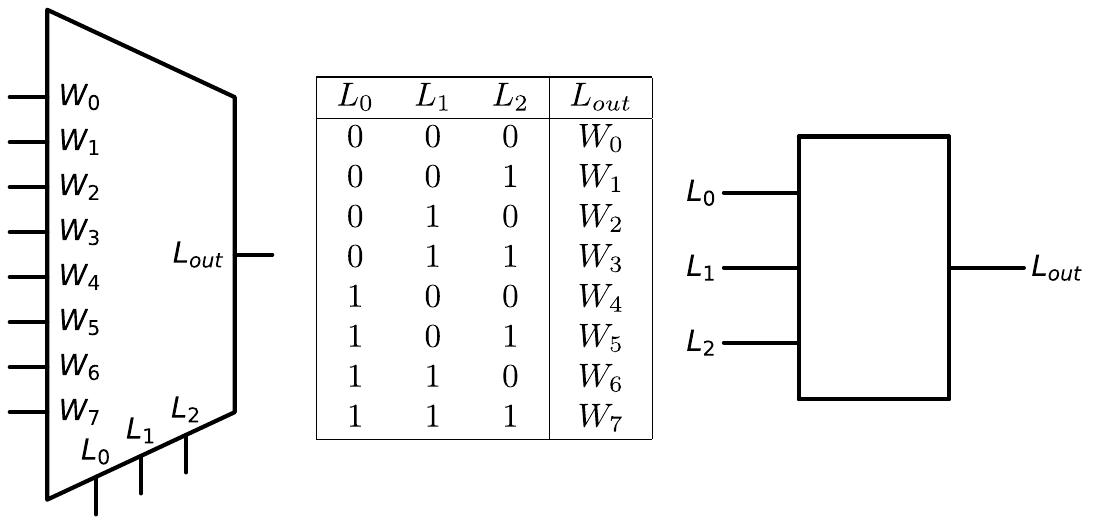}
     \caption{Equivalence between an 8:1 MUX and a 3-input LUT. Left: An 8:1 MUX with 8 input values, 3 selector bits and one output value. Middle: A lookup table with entries $W_0,\ldots,W_7$, which are input values of the MUX. Right: The symbol we will employ for a LUT with three input values and one output value.}
     \label{fig:mux_lut}
    \end{figure}

    Here $L_{out}$ is the output value of the MUX, $W_0,\ldots, W_7$ are the input values of the MUX, and $L_0, L_1$ and $L_2$ are the selector bits. Alternatively, this can also be viewed as a 3-input LUT, where the input values $L_0, L_1$, and $L_2$ determine which of the eight LUT entries $W_0, \ldots, W_7$ is selected to produce the output value $L_{out}$. The equivalence between a LUT and a MUX is shown graphically in Fig. \ref{fig:mux_lut}. Note that this formula is differentiable; hence, backpropagation can be used to find the optimal LUT entries. During training, full-precision values between 0 and 1 are propagated through the network. At inference time, the network only utilizes binary values.\\
    
    In general, an $N$-input LUT is represented by a $2^N$:$1$ MUX. The generalized formula for an $N$-input LUT is given by
    
    \begin{align}
        L_{\mathrm{out}} &= \sum_{i=0}^{2^N-1} W_i \prod_{j=0}^{N-1} \left( \bar{s}_{i,j}\bar{L}_j+s_{i,j}L_j \right) \label{eq:LUT_eq} \\   
        \text{with} \quad \mathbf{s} &= \begin{pmatrix}
            0 & 0 & \cdots & 0 & 0 \\ 
            0 & 0 & \cdots & 0 & 1 \\ 
            0 & 0 & \cdots & 1 & 0 \\ 
            & & \vdots & & \\
            1 & 1 & \cdots & 1 & 1 \label{eq:s_matrix}.\\
        \end{pmatrix}.
    \end{align}
    
     Here, $W_i$ are the trainable weights, which are the entries of the LUT. The values of $W_i$ are continuous and are binarized during training as in \cite{keuninckx_training_2025}, but in a layer-wise fashion. The sum in this equation is performed over all $2^N$ LUT-entries, where $N$ is the number of input pins of the LUT. In addition, $L_j$ with $j=0, \ldots,N-1$ are the input values of the LUT. $L_{out}$ is the LUT output value, which is the LUT entry that is selected based on the input values $L_j$. In this general formula, the constant $2^N\times N$  selector matrix $\mathbf{s}$, which lists integers 0 through $2^N-1$ in binary format, is used to select the correct Boolean variables $\overline{L_j}$ or $L_j$ for each input index $j$ and each LUT entry index $i$.\\

     During training, one must consider the batch size and the number of LUTs per layer, so that the above $L_{out}$ variable becomes a matrix of size $B\times O$, with $B$ the batch size and $O$ the layer size. The PyTorch pseudo-code for training a layer of LUTs that makes use of the above method is presented in Algorithm \ref{alg:Forward_LUTN}. Here $B$ is the batch size, $I$ is the input size of a layer, $O$ is the output size of a layer, and $N$ is the number of input pins of a LUT.
    
    \begin{algorithm}
    \caption{Forward pass of a LUTN layer}
    \label{alg:Forward_LUTN}
        \begin{algorithmic}
        \REQUIRE $dim(input) = B \times I$
        \REQUIRE $dim(idx) = O \times N$
        \REQUIRE $dim(W) = 2^N \times O$
        \REQUIRE $dim(s) = 2^N \times N$
        \ENSURE $L_{out}$ is calculated according to Eq. \eqref{eq:LUT_eq}
        \STATE $L \gets input[:,idx].unsqueeze(2)$
        \STATE $L \gets broadcast\_to(L,(B,O,2^N,N))$
        \STATE $L \gets torch.prod(L\cdot s + (1-L) (1-s),dim=-1)$
        \STATE $L \gets torch.transpose(L,1,2)$
        \STATE $L_{out} \gets torch.sum(L\cdot W,dim=1)$
        \RETURN $L_{out}$
        \end{algorithmic}
    \end{algorithm}
    
    \subsubsection{Probabilistic interpretation}
     
     While training, the $L_j$ values will be real numbers, and only after binarization will $L_j\in\{0,1\}$. We will now show that when these continuous valued $L_j$ parameters are employed in the aforementioned LUT equation, this will lead to a probabilistic interpretation of the LUT primitive. \\
     
     Let $X_j \in\{0,1\}$ be a Bernoulli distributed random variable with parameter $L_j\in[0,1]$. In addition, let $\mathbf{b}\in\{0,1\}^N$ be a LUT address, i.e. $b_j=s_{ij}$, where $N$ is the number of LUT input pins. Then the LUT-equation Eq. \eqref{eq:LUT_eq}, can be rewritten as  

    \begin{align}
        L_{\mathrm{out}} &= \sum_{\mathbf{b}\in\{0,1\}^{N}} W_{\mathbf{b}} \prod_{j=0}^{N-1} L_j^{b_j}(1-L_j)^{1-b_j} \\
        &= \sum_{\mathbf{b}\in\{0,1\}^{N}} W_{\mathbf{b}} \prod_{j=0}^{N-1} P(X_j=b_j) \\
        &= \sum_{\mathbf{b}\in\{0,1\}^{N}} W_{\mathbf{b}}\, P(\mathbf{X}=\mathbf{b}) \\
        &= \mathbb{E}\!\left[W_{\mathbf{X}}\right].
    \end{align}
    
       The input values of the LUT, $L_j\in[0,1]$, now represent the probability of the input being equal to one: $P(X_j=1)=L_j$. Note that we assumed independent probabilities, i.e. $P(\mathbf{X}=\mathbf{b})=P(X_0=b_0,\ldots,X_{N-1}=b_{N-1})=\prod_{j=0}^{N-1} P(X_j=b_j) $, and thus independent bit streams and rates $L_j$. In the last step, we defined a random variable $W_\mathbf{X}$, such that the LUT output value becomes equal to the expected value over all LUT entries. If $W_i\in[0,1]$, the LUT output value $L_{out}\in[0,1]$ represents the output rate of the LUT.
    
    \subsubsection{Relation to DWNs and WARP-LUTs}
    The proposed method and Differentiable Weightless Neural Networks (DWNs) \cite{bacellar_differentiable_2024} are closely related. Both methods directly train networks of differentiable $N$-input LUTs using backpropagation. When $N=6$, such LUTs naturally correspond to the primitive 6-LUTs available in modern FPGAs. To learn the LUT entries, DWNs employ an Extended Finite Difference method, while our proposed method employs a hardware-inspired differentiable MUX equation. Hence, both methods are able to learn the same LUT functions, but use a different differentiable LUT model during training. Our method places DWNs in a probabilistic interpretable setting and is purely used as an alternative to DWNs.\\

    The LUTNs also differ from WARP-LUTs \cite{gerlach_warp-luts_2025}, which parameterize LUTs using Walsh-Hadamard coefficients, reducing the number of learnable parameters compared to softmax-based LGNs. However, their method has only been evaluated for two-input LUTs. In addition, other LUT training methods such as LUTNet \cite{wang_lutnet_2019,wang_lutnet_2020}, LogicNets \cite{umuroglu_logicnets_2020}, NullaNet \cite{nazemi_energy-efficient_2019}, PolyLUT \cite{andronic_polylut_2023} and NeuraLUT \cite{andronic_neuralut_2024} do not learn LUT entries directly. Rather, they first train a neural network and subsequently map it to a LUT-based network.
    
    \subsection{Evaluation metrics} 
    Accuracy may not be the best evaluation metric for the MIT-BIH data set, as 88.58\% of the train set and 89.81\% of the test set data examples are normal heartbeats. This class imbalance means that any classifier that predicts every heartbeat as normal already achieves an accuracy of 89.81\%. Thus, we employ per-class evaluations of the precision or positive predictive value (P), sensitivity, recall or true positive rate (Se), specificity or true negative rate (Sp) and $F_1$ score. These are given by:

    \begin{align*}
        P &= \frac{TP}{TP+FP},\\
        Se &= \frac{TP}{TP+FN}, \\
        Sp &= \frac{TN}{TN+FP}, \\
        F_1\text{ score}&=\frac{2}{P^{-1}+Se^{-1}}.
    \end{align*} 
    
    Here $TP$, $FP$, $TN$, $FN$ are the true positives, false positives, true negatives and false negatives respectively. In addition, we report the $j\kappa$ index which has been proposed to quantify the performance on the most important arrhythmia types \cite{mar_optimization_2011,mondejar-guerra_heartbeat_2019} better, namely the supraventricular ectopic beats (S) and ventricular ectopic beats (V) by employing the $j$ index. It also takes into account the class imbalance through the $\kappa$ index. The $j\kappa$ index consists of the $j$ index and the Cohen's Kappa index $\kappa$ as follows:
    
    \begin{equation}
        j\kappa = 1/8\cdot \,j + 1/2 \cdot \kappa.
    \end{equation}
    
      To make sure that $j\kappa$ $\in[0,1]$, coefficients of 1/8 and 1/2 are chosen, since $j$ $\in[0,4]$ and $\kappa$ $\in[0,1]$. The $j$ index is given by \cite{mar_optimization_2011}:

      \begin{equation}
          j = \text{Se\textsubscript{S}} + \text{Se\textsubscript{V}} + \text{P\textsubscript{S}} + \text{P\textsubscript{V}},
      \end{equation}
    and quantifies the classifier's performance of the S class and V class based on sensitivity and precision. The $\kappa$ index is given by \cite{fatourechi_comparison_2008,mondejar-guerra_heartbeat_2019}
    \begin{equation}
        \kappa = \frac{Acc-p_e}{1-p_e},
    \end{equation}
    where $Acc$ is the overall accuracy and $p_e$ is the chance agreement \cite{fatourechi_comparison_2008,mondejar-guerra_heartbeat_2019}:

    \begin{align*}    
        p_e &= \sum_k P(T=k)P(Pr=k)\\
        &=\frac{\sum_k\left[\sum_iC_{ik}\cdot\sum_jC_{kj}\right]}{\left(\sum_{ij} C_{ij}\right)^2}.
    \end{align*}
    Here, $P(T=k)P(Pr=k)$ is the probability that both the true label and predicted label are of class $k$, assuming independence $P(T=k,Pr=k)=P(T=k)P(Pr=k)$. Summing over all classes yields the probability to obtain a correct classification by chance alone, given by $p_e$. These probabilities can be obtained from the confusion matrix $C$. The chance agreement $p_e$ is compared with the actual accuracy of the classifier and normalized to obtain Cohen's Kappa. Therefore, the $\kappa$ index is a more robust metric for imbalanced data sets compared to just the overall or mean accuracy \cite{fatourechi_comparison_2008,mondejar-guerra_heartbeat_2019}. For example, assume a classifier that classifies normal heartbeats (N) and three different arrhythmias (S, V, F). Assume that we have 100 examples, 90 of which are normal beats, 5 of class S, 4 of class V and 1 of class F. Assume that all heartbeats are classified as normal, so that we have a high accuracy of 90\%. In that case, we obtain $p_e=0.9$ and $\kappa=0$, showing that even with a high accuracy the $\kappa$ value is lowest when all beats are classified as normal heartbeats.\\

    \subsection{Experimental protocol}
    Before benchmarking the LGNs and LUTNs on the MIT-BIH data set, preliminary experiments were conducted on the MNIST Handwritten Digits and Fashion-MNIST data sets to assess the validity of rate coding and the LUTN training algorithm. The findings are presented in \ref{sec:validation}. Next, the ECG arrhythmia data set is benchmarked employing both the mixed-patient and inter-patient paradigm. In the latter case, models are evaluated utilizing all of the preceding metrics, alongside an estimate of the number of FLOPs. Additionally, the design is implemented on a Xilinx Zynq-7000 ZedBoard, providing resource, power and latency estimations in a practical setting.

\section{Experimental Results}
This section first validates rate coding of the LGNs. The performance of the LGNs with rate-coded inputs are presented in \ref{sec:validation_ratecoding} on the MNIST Handwritten Digits and Fashion-MNIST data sets. The results showcase that using rate coding results in a higher accuracy on the MNIST Handwritten Digits and Fashion-MNIST data sets compared to threshold coding. In addition, the proposed LUTN training method is validated on the same data sets in \ref{sec:validation_LUTN}, demonstrating a higher accuracy compared to LGNs in all cases. We find that networks using rate-coded inputs always outperform networks without rate-coded inputs, and networks of LUTs with a greater fan-in outperform LUT networks with a lower fan-in. The remainder of this section is dedicated to ECG arrhythmia classification by benchmarking trained LGNs and LUTNs on the MIT-BIH data set. First, the mixed-patient paradigm is employed to train LGNs. The performance of these trained LGNs is compared to state-of-the-art (SOTA) LGN results to assess how well our preprocessing method improves their performance. Next, the inter-patient paradigm is utilized to analyze how well our classifiers can generalize to unseen patient data. Our work is compared to SOTA methods based on accuracy, $j\kappa$ index, per-class metrics and number of FLOPs.

\subsection{Evaluation on the MIT-BIH data set} \label{sec:MIT_BIH}
\subsubsection{Mixed-patient paradigm}
To evaluate our preprocessing methodology, we trained \linebreak[4] LGNs using our preprocessing method and compared them to SOTA LGN results. Table \ref{tab:Mixed_patients} gives the accuracy on the MIT-BIH mixed-patient test set of our preprocessing method in comparison with the SOTA method that only utilizes dynamic thresholds. All networks consisted of 8000 gates per layer, and were trained for 200 epochs with a batch size of 100 examples, using the same randomly chosen train-test split of 0.67 -- 0.33 for all models. As mentioned in Section \ref{sec:preprocessing}, four classes were utilized in all of our models, since the fifth class only consists of 15 examples. Feng et al. \cite{feng_low-power_2024} report accuracies of 97.54\%, 97.86\% and 98.05\% for a 2-layer, 3-layer and 4-layer network respectively. However, under previously mentioned training circumstances of 200 epochs and a batch size of 100 examples, the accuracies are equal to 97.30\%, 97.44\% and 97.47\%, as seen in Table \ref{tab:Mixed_patients}. Our preprocessing method significantly increased the accuracy on the MIT-BIH data set for the mixed-patient case. Table \ref{tab:Mixed_patients_temp} in \ref{sec:optimal_temp} gives the optimal temperature of the applied softmax function to the population counters during training.\\

\begin{table}[t]
    \centering
    \caption{Accuracy (\%) on the mixed-patient MIT-BIH benchmark using LGN classifiers with varying number of layers. The second and third rows use the binary and full-precision features described in this work respectively. The superscript $r$ indicates the use of rate coding. The first row only uses dynamic thresholds as preprocessing \cite{feng_low-power_2024}.}
    \label{tab:Mixed_patients}
    \begin{tabular}{lllll}
    \toprule
    Models & Ref. & 2 layer & 3 layers & 4 layers \\
    \midrule
    LGNs & \cite{feng_low-power_2024} & 97.30 & 97.44 & 97.47 \\
    \textbf{LGNs} & \textbf{Proposed} & \textbf{97.90} & \textbf{98.25} & \textbf{98.35} \\
    LGNs$^r$ & Proposed & 97.56 & 97.64 & 97.76 \\
    \bottomrule
    \end{tabular}
\end{table}

\subsubsection{Inter-patient paradigm} \label{sec:inter_patient}
\begin{table}[t] 
\centering
\caption{Comparison of our trained models with state-of-the-art results on the MIT-BIH data set employing the inter-patient paradigm. For our models, the best $j\kappa$ index and accuracy values across all evaluated network depths are reported. A dash means that this metric could not be determined from the reference.}
\label{tab:inter_patient_summary}
\resizebox{\columnwidth}{!}{%
\begin{tabular}{lccccc}
\toprule Models & Ref. & $j$ index & $\kappa$ index & $j\kappa$ index & Acc. (\%) \\
\midrule
LGN & \cite{feng_low-power_2024} & 1.540 & 0.483 & 0.434 & 91.13 \\
SNN & \cite{mao_ultra-energy-efficient_2022} & -- & -- & -- & 93.70 \\
SNN$^r$ & \cite{yan_energy_2021} & 2.490 & 0.595 & 0.609 & 90.00 \\
SNN$^r$+CNN & \cite{yan_energy_2021} & 2.660 & 0.597 & 0.631 & 90.00 \\
Deep CNN & \cite{li_inter-patient_2022} & 2.693 & 0.729 & 0.701 & 89.00 \\
SVM Ensemble & \cite{mondejar-guerra_heartbeat_2019} & 3.164 & 0.755 & \textbf{0.773} & \textbf{94.50} \\
LS-SVM & \cite{villa_are_2019} & 2.830 & 0.500 & 0.610 & 81.00 \\
\midrule
LGN & Proposed & 2.576 & 0.657 & 0.650 & 94.28 \\
LGN$^r$ & Proposed & 2.768 & 0.675 & \textbf{0.683} & 93.63 \\
2-LUTN & Proposed & 2.693 & 0.666 & 0.669 & \textbf{94.41} \\
4-LUTN & Proposed & 2.634 & 0.633 & 0.646 & 94.26 \\
6-LUTN & Proposed & 2.661 & 0.636 & 0.651 & 94.24 \\
\bottomrule
\end{tabular}}
\end{table}

\begin{table*}[t]
\centering
\caption{The per-class precision (P), sensitivity (Se), specificity (Sp) and F$_1$-score of our trained LGNs and LUTNs compared to other works on the MIT-BIH inter-patient test set. The superscript $r$ indicates a rate-coded network. The blank entries represent values that could not be determined in a reliable manner.}
\label{tab:perclass_metrics}

\setlength{\tabcolsep}{2.0pt} 
\renewcommand{\arraystretch}{0.85} 
\fontsize{6}{7}\selectfont 

\resizebox{\textwidth}{!}{%
\begin{tabular}{lc|cccc|cccc|cccc|cccc}
\toprule

& &
\multicolumn{4}{c|}{\textbf{N}} & 
\multicolumn{4}{c|}{\textbf{S}} & 
\multicolumn{4}{c|}{\textbf{V}} & 
\multicolumn{4}{c}{\textbf{F}} \\ 
\cmidrule(lr){3-6} 
\cmidrule(lr){7-10} 
\cmidrule(lr){11-14} 
\cmidrule(lr){15-18} 

Models & Ref. &
P & Se & Sp & $F_1$ & 
P & Se & Sp & $F_1$ & 
P & Se & Sp & $F_1$ & 
P & Se & Sp & $F_1$ \\ 

\midrule 

SNN$^r$ & \cite{yan_energy_2021} & 0.970 & 0.920 & 0.779 & 0.944 & 0.460 & 0.670 & 0.966 & 0.545 & 0.590 & 0.770 & 0.963 & 0.668 & -- & -- & -- & --  \\
SNN$^r$ + CNN & \cite{yan_energy_2021} & 0.970 & 0.920 & 0.779 & 0.944 & 0.360 & 0.680 & 0.948 & 0.471 & 0.840 & 0.780 & 0.990 & 0.809 & -- & -- & -- & -- \\
Deep CNN & \cite{li_inter-patient_2022} & 0.933 & 0.945 & 0.808 & 0.939 & 0.798 & 0.884 & 0.949 & 0.839 & 0.659 & 0.352 & 0.988 & 0.459 & 0.450 & 0.424 & 0.992 & 0.437 \\ 
SVM Ensemble & \cite{mondejar-guerra_heartbeat_2019} & 0.982 & 0.959 & 0.863 & 0.971 & 0.498 & 0.781 & 0.966 & 0.608 & 0.938 & 0.948 & 0.996 & 0.943 & 0.236 & 0.124 & 0.997 & 0.162\\ 
LS-SVM & \cite{villa_are_2019} & 0.980 & 0.820 & 0.864 & 0.893 & 0.290 & 0.830 & 0.922 & 0.430 & 0.840 & 0.880 & 0.988 & 0.860 & 0.080 & 0.110 & 0.990 & 0.093 \\
 & & & & & & & & & & & & & & & & & \\
\hline
 & & & & & & & & & & & & & & & & & \\
LGN & Proposed & 0.946 & 0.988 & 0.558 & 0.966 & 0.689 & 0.127 & 0.998 & 0.214 & 0.906 & 0.867 & 0.994 & 0.886 & 0.022 & 0.013 & 0.995 & 0.016\\
LGN$^r$ & Proposed & 0.958 & 0.972 & 0.667 & 0.965 & 0.745 & 0.338 & 0.995 & 0.465 & 0.795 & 0.908 & 0.984 & 0.848 & 0.013 & 0.013 & 0.992 & 0.013\\ 
2-LUTN & Proposed & 0.945 & 0.997 & 0.546 & 0.970 & 0.793 & 0.047 & 0.999 & 0.089 & 0.953 & 0.901 & 0.997 & 0.926 & 0.059 & 0.005 & 0.999 & 0.009 \\
4-LUTN & Proposed & 0.940 & 0.997 & 0.506 & 0.968 & 0.810 & 0.008 & 1.000 & 0.016 & 0.951 & 0.865 & 0.997 & 0.906 & 0.000 & 0.000 & 0.999 & 0.000\\
6-LUTN & Proposed & 0.941 & 0.998 & 0.511 & 0.968 & 0.833 & 0.007 & 1.000 & 0.015 & 0.952 & 0.869 & 0.997 & 0.908 & 0.023 & 0.003 & 0.999 & 0.005\\ 
\bottomrule 
\end{tabular}%
} \end{table*}

To assess the performance of the LGNs and LUTNs for ECG arrhythmia classification on unseen patients, both the \linebreak[4] LGNs and LUTNs were evaluated on the MIT-BIH data set using the inter-patient paradigm. The performance of the LGNs and LUTNs is presented in Table \ref{tab:inter_patient_summary}. The LGNs/2-LUTNs, 4-LUTNs, and 6-LUTNs use 8000 gates/LUTs per layer (16000 connections per layer), 3000 LUTs per layer (12000 connections per layer), and 2000 LUTs per layer (12000 connections per layer) respectively. Both binary features and numerical features (rate-coded) were tested. In addition, our results are compared to those of the LGNs in \cite{feng_low-power_2024}, which used only dynamic thresholds for preprocessing. From these results, it can be concluded that using just thresholds is not enough for performing well on this data set, e.g. the normal class occurs around 89\% of the time, which should be the lower bound of the accuracy, and the networks of \cite{feng_low-power_2024} obtain 91.13\%. In addition, the $\kappa$ index, which is a metric that takes into account this imbalance, equals 0.483. Our models obtain an accuracy above 94\%, with a $\kappa$ index up to 0.675. The $j$ index, which takes into account the sensitivity and precision of the most important arrhythmias is equal to 1.540 with dynamic threshold encoding, while the networks employing our preprocessing achieve a value between 2.576 and 2.768. These results show that a proper preprocessing method is of vital importance to generalize well to unseen patient data.\\

Table \ref{tab:inter_patient_summary} also compares our work with an SNN \cite{mao_ultra-energy-efficient_2022}, a rate-coded SNN (SNN$^r$) \cite{yan_energy_2021}, a rate-coded SNN in combination with a CNN \cite{yan_energy_2021}, a deep CNN \cite{li_inter-patient_2022}, an ensemble of SVMs \cite{mondejar-guerra_heartbeat_2019} and an LS-SVM \cite{villa_are_2019}. Compared to these SOTA results, our models attain the highest accuracy of 94.41\%, except for an ensemble of SVMs (94.50\%). Based on the $j\kappa$ index, only a deep CNN ($j\kappa=0.701$) and an ensemble of SVMs ($j\kappa=0.773$) perform better than our models ($j\kappa=0.683$). Nevertheless, these SOTA models do not take into account the energy usage, which will be discussed in Section \ref{sec:FLOPs_estimation}. The optimal softmax function temperatures that are applied to the population counters for our models employing the inter-patient paradigm are shown in Table \ref{tab:temp_inter_patient} of \ref{sec:optimal_temp}. More detailed results about the accuracy and $j\kappa$ index are presented in Table \ref{tab:Acc_inter_patient}, Table \ref{tab:jk_inter_patient} and Table \ref{tab:clamping} of \ref{sec:detailed_results}. We found that using binarized input features enhances the performance of the classifiers. To test this, a total of 60 multilayer perceptrons (MLPs) were trained on both continuous-valued and binary input data with different seeds. Utilizing binary features, we obtained an accuracy of $94.11\pm0.18\%$ and a $j\kappa$ index of $0.669\pm0.012$, while for continuous-valued inputs we achieved an accuracy of $93.09\pm 0.32\%$ and a $j\kappa$ index of $0.608\pm 0.050$. From the tested models (binary input MLP, MLP, LUTN, LUTN$^r$, LGN, LGN$^r$), only LGN$^r$ obtains a higher $j\kappa$ index on average compared to its binary equivalent. In addition, to have stable training of high-fan-in LUTs, clamping weights between zero and one seemed to be crucial. Note that rate coding is performed with full-precision values, which corresponds to an infinite bit stream length. As a consequence, our reported performance relates to optimally encoded input values. In practice, a trade-off exists between performance and energy usage: A shorter bit stream duration results in a higher performance degradation, as shown in Fig. \ref{fig:acc_jk_rate}, while reducing the required energy per inference.\\

Table \ref{tab:perclass_metrics} shows the per-class precision, sensitivity, specificity and $F_1$-score for our methods compared to the SOTA. Our models tend to have slightly lower specificity for the normal-beat class than less-energy-efficient SOTA models. The LGNs and LUTNs obtained a higher precision for the supraventricular ectopic (S) beats compared to other works, meaning that if an S beat is predicted, it is almost certainly an S beat. In turn this means that the sensitivity is lower compared to other works, implying that many S beats are classified as normal beats. The LGN with rate-coded inputs (LGN$^r$) is our most suitable model for predicting the S class, having a precision of 0.745, a sensitivity of 0.338 and an $F_1$-score of 0.465. The performance of our models on ventricular ectopic (V) beats show that they achieve a higher precision (up to 0.952) and higher sensitivity (up to 0.908), resulting in a higher $F_1$-score (up to 0.908) compared to the SOTA. Only the deep CNN achieves a higher $F_1$-score of 0.943. Lastly, the F class is very hard to detect due to the low number of examples, and again, the deep CNN has the highest $F_1$-score of 0.437. Table \ref{tab:confusion_matrix} displays the confusion matrix for the rate-coded LGN, and confirms our previous per-class reasoning. Both the normal and V beats are classified very well, while a significant number of S and F beats are classified as normal beats. Of the 43761 normal beats, only 1204 were classified as arrhythmias. Nevertheless, compared to the very low F class counts, the 355 normal beats classified as F beats have a large impact on the F class precision, shown in Table \ref{tab:perclass_metrics}. Many F class examples (383), compared to the total number of F class examples (388), are confused as normal (289) or ventricular (93) ectopic beats, resulting in a low sensitivity in Table \ref{tab:perclass_metrics}.\\

\begin{table}[t]
\centering
\caption{Confusion matrix of our 2-layer LGN with rate-coded inputs on the MIT-BIH inter-patient test set.}
\label{tab:confusion_matrix}
\begin{tabular}{lllll} 
\hline
 & \textbf{N} & \textbf{S} & \textbf{V} & \textbf{F} \\
\hline
\textbf{N} & 42557 & 222 & 627 & 355 \\
\textbf{S} & 1317 & 691 & 32 & 3 \\
\textbf{V} & 269 & 14 & 2913 & 12 \\
\textbf{F} & 289 & 1 & 93 & 5 \\
\hline
\end{tabular}
\end{table}

\subsubsection{Estimating the number of FLOPs} \label{sec:FLOPs_estimation}

\begin{table*}[h]
    \centering
    \caption{Comparison of the number of floating-point operations per inference (FLOPs) for different approaches. The total number of FLOPs is the sum of the FLOPs for preprocessing, the network, and the readout. Only the convolution operations are counted for the ANN network FLOPs. Our networks presented here are only 1-layer feedforward architectures of 8000 gates/2-LUTs, 3000 4-LUTs or 2000 6-LUTs. Still, the number of network FLOPs can be extended to multilayer networks by multiplying the current value with the number of layers. In that case, the number of FLOPs for the preprocessing and readout stay the same. $^1$Not computed, as the reported value is already notably higher than those of the proposed methods, and would not affect the comparative conclusions.}
    \label{tab:flops}
    \begin{tabular}{lllllll}
    \toprule
    Models & Ref. & Architecture & Preprocessing & Network & Readout & Total \\
     & & & FLOPs & FLOPs & FLOPs & FLOPs \\
    \midrule
    ANN & \cite{li_inter-patient_2022} & Conv. & Not computed$^1$ & $1.35 \cdot 10^9$ & 0 & $>$ 1.35 G \\ 
SNN$^r$ & \cite{yan_energy_2021} & Conv. & 0 & $31.0 \cdot 10^6$ & Not computed$^1$  & $>$ 31.0 M \\
    SVM & \cite{mondejar-guerra_heartbeat_2019} & Ensemble & Not computed$^1$ & $4.93 \cdot 10^6$ & 0 & $>$ 4.93 M \\
    LGNs & Proposed & 1x8000 & 2246 & \textbf{80} & 560 & \textbf{2.89 K} \\
    2-LUTN & Proposed & 1x8000 & 2246 & 720 & 560 & 3.53 K \\
    4-LUTN & Proposed & 1x3000 & 2246 & 1350 & 211 & 3.81 K \\
    6-LUTN & Proposed & 1x2000 & 2246 & 3780 & \textbf{139} & 6.17 K \\
    \bottomrule
    \end{tabular}
\end{table*}

With the goal of determining the energy expenditure of our networks, and comparing it to state-of-the-art literature, we estimate the number of floating-point operations (FLOPs) per inference as shown in Table \ref{tab:flops}. To estimate the number of FLOPs for the LGNs, we use the conservative estimate of 100 binary operations (i.e. gate operations) for one floating-point operation as mentioned in original LGN work \cite{petersen_deep_2022}. In reality, it is expected that one FLOP requires much more than 100 gate operations, and as such, the number of FLOPs for the LGNs (and by extension the LUTNs) is most likely underestimated. The number of FLOPs for the LUTNs is calculated in the same manner, taking into account that an N-LUT is represented by a 2$^N$:1 MUX that contains $3(2^N-1)$ gates. To estimate the number of FLOPs of the readout, the number of gates of a ripple carry adder tree is counted and converted to FLOPs. Table \ref{tab:flops} displays the results for the top-performing SOTA models and a rate-coded SNN. Although the SVM achieves the best performance in terms of accuracy and $j\kappa$ index in Table \ref{tab:inter_patient_summary}, the LGNs and LUTNs require about three orders of magnitude less FLOPs. As mentioned, it is expected that the number of FLOPs for the LGNs and LUTNs are significantly lower, due to our conservative conversion rate from gates to FLOPs. Additionally, the number of FLOPs for the preprocessing is mainly dominated by the calculation of the crest factor, which is discussed in more detail in the next section.

\section{Hardware Implementation and Evaluation}

    \subsection{System architecture}
\begin{figure*}[h!]
        \centering
         \includegraphics[width=\textwidth,keepaspectratio]{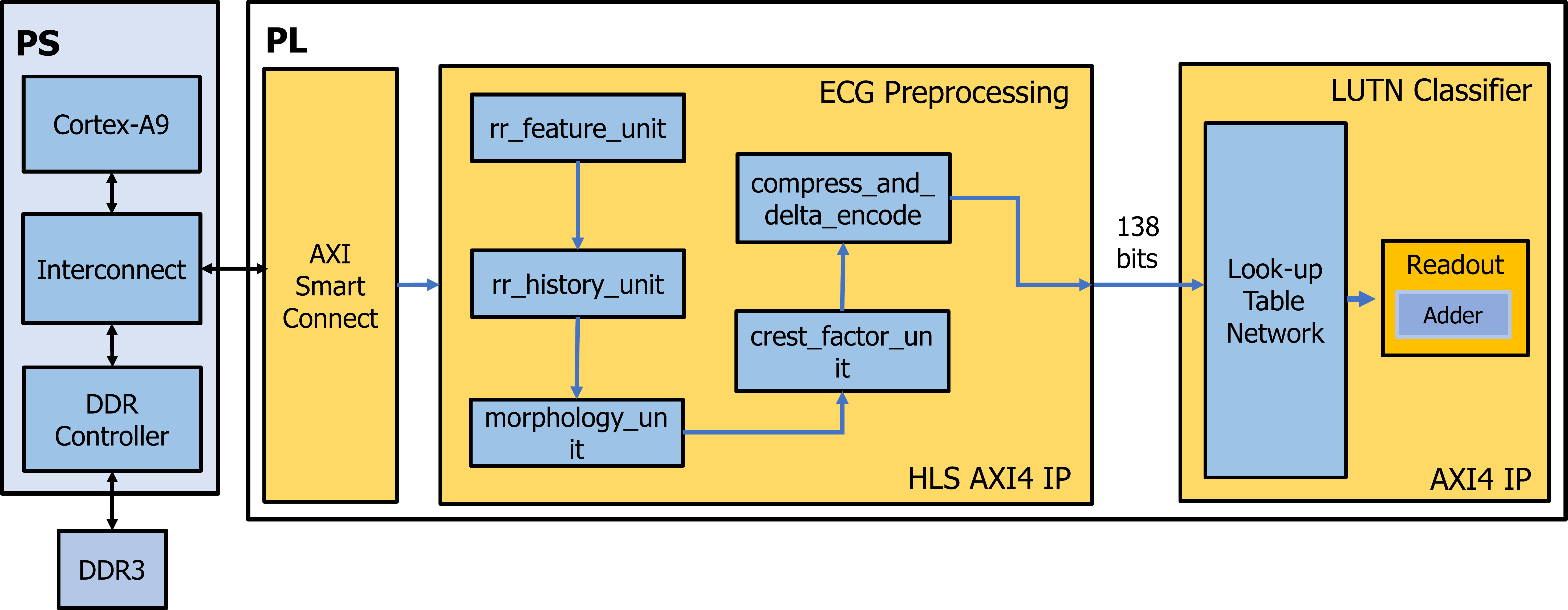}
         \caption{The block diagram of the full ECG preprocessing and LUTN classification pipeline, implemented on a Xilinx Zynq-7000 ZedBoard. The Processing System (PS) includes the dual-core Arm Cortex-A9 processor, memory controllers and AXI interconnect interfaces. Through this interconnect (i.e. AXI SmartConnect), the PS can control the custom IPs in the Programmable Logic (PL). The ECG processing IP reads raw ECG data and extracts the wanted binary features as explained in Section \ref{sec:feature_extraction}. The 138-bit wide feature vector is fed into the single layer LUTN consisting of 2000 6-LUTs, after which one 9-bit adder is reused four times to calculate the class scores. The class with the highest score represents predicted class label.}
         \label{fig:block_diagram}
    \end{figure*}

    To assess the deployment of LUTN-based arrhythmia classification, the full ECG preprocessing and LUTN classification pipeline was implemented on an Xilinx Zynq-7000 ZedBoard (part number xc7z020clg484-1). We chose to implement a LUTN instead of an LGN, given that the LUTN is inherently reprogrammable on both ASICs and FPGAs, such that if additional heartbeat data is available, the network can be retrained and new lookup table entries can be uploaded. The FPGA in our design uses 6-LUTs as the core elements of its reconfigurable logic blocks, making a 6-LUTN the ideal model choice to implement. The block diagram is shown in Fig. \ref{fig:block_diagram}, displaying the Processing System (PS) and Programmable Logic (PL). The PS consists of a dual-core Arm Cortex-A9 processor, memory interfaces, I/O peripherals and AXI interface ports. The latter enables communication between the ARM processor and custom IP cores on the PL through AXI SmartConnect. In this case, the custom IP cores are the ECG Preprocessing IP for heartbeat preprocessing and LUTN IP for heartbeat classification. We employed Vitis HLS (version 2025.1) to convert the C++ preprocessing code of Section \ref{sec:feature_extraction} to an ECG Preprocessing IP. A Python script was used that automatically converted the trained LUTN along with readout into Verilog files, creating the LUTN Classifier IP. The next two sections explain each of these IPs in more detail.
    
    \begin{table*}[h]
    \centering
    \caption{The number of lookup tables (LUTs), flip-flops (FF), BRAMs and DSPs provided by the post-implementation report in Vivado for each hardware block and its components. The power is provided from the post-implementation report and converted to energy per inference employing a clock frequency of 1 MHz. The latency of the ECG Preprocessing IP is determined from the Vitis C/RTL Cosimulation Timeline Trace, while the latency of the LUTN Classifier IP is determined from Vivado. $^1$Only combinational logic. $^2$Control logic adds no latency.} \label{tab:hardware_results}
        \begin{tabular}{@{}lrrrrrrrr@{}}
        \toprule
        \textbf{Hardware Block} & \textbf{LUTs} & \textbf{FFs} & \textbf{BRAMs} & \textbf{DSPs} & \textbf{Latency} & \textbf{Latency} & \textbf{Dynamic Power} & \textbf{Energy} \\
        & & & & & \textbf{(clk cycles)} & \textbf{(ms)} & \textbf{(mW)} & \textbf{($\mu$J)} \\
        \midrule \textbf{ECG Preprocessing IP} & \textbf{17064} & \textbf{2811} & \textbf{3.5} & \textbf{53} & \textbf{3540} & \textbf{3.540} & \textbf{2.33} & \textbf{8.25} \\
        \quad \texttt{rr\_feature\_unit} & 3697 & 0 & 0 & 20 & 0$^1$ & 0$^1$ & 0.42 & 1.49 \\ \quad
        \texttt{rr\_history\_unit} & 3194 & 314 & 0 & 9 & 504 & 0.504 & 0.59 & 2.09 \\
        \quad \texttt{morphology\_unit} & 2903 & 735 & 0 & 3 & 295 & 0.295 & 0.24 & 0.85 \\ \quad
        \texttt{crest\_factor\_unit} & 6254 & 574 & 0 & 21 & 1754 & 1.754 & 0.99 & 3.50 \\
        \quad \texttt{compress\_and\_delta\_encode} & 319 & 267 & 0.5 & 0 & 478 & 0.478 & 0.02 & 0.07 \\
        \quad Interface \& control logic & 697 & 921 & 3 & 0 & 509 & 0.509 & 0.07 & 0.25 \\ \addlinespace
        \textbf{LUTN Classifier IP} & \textbf{3223} & \textbf{267} & \textbf{0} & \textbf{0} & \textbf{5} & \textbf{0.005} & \textbf{0.092} & \textbf{0.00046} \\ 
        \quad LUTN & 2000 & 0 & 0 & 0 & 1 & 0.001 & 0.004 & 0.00002 \\
        \quad Popcount/Adder & 1122 & 0 & 0 & 0 & 4 & 0.004 & 0.070 & 0.00035 \\ 
        \quad Interface \& control logic & 101 & 267 & 0 & 0 & 0$^2$ & 0$^2$ & 0.018 & 0.00009 \\
        \bottomrule
    \end{tabular}
    \end{table*}

    \subsection{ECG preprocessing accelerator} 
    The ECG Preprocessing IP consists of five RTL modules, each of them representing one of the preprocessing steps in Section \ref{sec:feature_extraction}. First, the \texttt{rr\_feature\_unit} extracts the RR-intervals $RR_1$, $RR_2$, $RR_3$ and $RR_4$, encoding each of them into 8 bits. It also determines the RR-interval changes $\Delta RR_p$ and $\Delta RR_m$, each representing one bit. Note that no peak detection algorithm is implemented. Instead, the peak annotations from the MIT-BIH data set are utilized. Next, the \texttt{rr\_history\_unit} is employed to determine $RR_{locCV}$ and $RR_{ratio}$, each of which are encoded into two bits. This module also determines the tachycardia bit. Following this, the \texttt{morphology\_unit} calculates morphology features $M_1$, $M_2$, $M_4$,  encoding each one into three bits. Hereafter, the \texttt{crest\_factor\_unit} determines both 8-bit crest factors $cf_1$ and $cf_2$. Lastly, the \texttt{compress\_and\_delta\_encode} module performs delta encoding, obtaining a 74-bit feature value. All these features are represented as one 138-bit long feature vector that is passed to the LUTN classifier.
    
    \subsection{LUTN classifier accelerator}
    The LUTN Classifier IP receives a 138-bit feature vector that is extracted from every heartbeat. The feature vector is used as input for the LUTN that consists of one layer of 2000 6-LUTs. At the end of the network, the output values of the LUTs in the final layer are added in chunks of 500 LUTs, where each chunk corresponds with one class population counter. Since the LUTN itself requires less than a clock cycle of latency and the ECG arrhythmia classification does not need such rapid predictions, we chose to reuse one 9-bit adder for each class label, instead of instantiating four adders in parallel. Each class label is associated with a sum, and the sum with the highest final value corresponds to the predicted class label.
    
    \subsection{Hardware integration and data flow}
    The peak positions along with the ECG sample values are given to the ECG Preprocessing IP through the AXI Smart Connect. After the ECG preprocessing is finished, the 138-bit feature vector is passed to the LUTN Classifier IP, where a one-layer LUTN consisting of 2000 6-LUTs performs the classification. This IP employs one adder that is used four times, once for each class score.
    
    \subsection{Resource utilization}
    Table \ref{tab:hardware_results} presents the number of LUTs, FFs, BRAMs and DSPs of both the ECG Preprocessing IP and the LUTN Classifier IP for a single layer LUTN consisting of 2000 6-LUTs. These numbers were obtained from the post-implementation Vivado (version 2025.1) report. The LUTN Classifier IP uses significantly fewer LUTs compared to the ECG Preprocessing IP, showcasing that the LUTN can be use as a hardware-efficient classifier add-on to the necessary ECG preprocessing. In detail, the LUTN itself employs 2000 LUTs and 0 FFs, while the readout utilizes 1122 LUTs and 0 FFs. The rest of the utilization stems from interface and control logic. Note that LUTN topology, i.e. the number of layers and the layer-size for each layer, was chosen in PyTorch as a hyperparameter before training. After the network is trained, it is automatically converted to Verilog files, and converted to a classifier IP. In this case, the chosen topology was one layer of 2000 6-LUTs. Based on the post-implementation report of Vivado, the LUTN is indeed mapped to exactly 2000 6-LUTs. This demonstrates that the topology selected prior to training remains valid in the final implementation, enabling a one-to-one software-to-hardware mapping.\\
    
    The ECG Preprocessing IP employs 17064 LUTs, 2811 FFs, 3.5 BRAMs and 53 DSPs. Most of those utilization numbers are used by the five modules. Again some LUTs, FFs and BRAMs are used as interface and control logic. The crest factors utilize most of the LUTs (6254), while the delta encoding requires the least number of LUTs (319) compared to the other modules.

    \subsection{Latency, power, and energy evaluation}
    Table \ref{tab:hardware_results} also shows the latency, power and energy estimation of both IPs for one inference, employing a clock frequency of 1~MHz. These results were obtained from the post-imple-\linebreak[4]mentation report in Vivado and from the Vitis C/RTL Cosimulation Timeline Trace. The Vivado report shows only the dynamic power usage for each IP and their subcomponents, employing a vectorless analysis that results in a medium confidence level. The ECG Preprocessing IP requires 3540 clock cycles, or 3.540 ms to complete, while the LUTN Classifier IP needs 5 clock cycles or 0.005 ms. This combined latency is well below the 273~ms inter-beat interval corresponding to a maximum heart rate of 220~bpm. On this Xilinx Zynq-7000 ZedBoard, the energy per inference estimate is equal to 8.25 µJ, of which almost all stems from the preprocessing IP. The full LUTN IP with popcount and interface/control only consumes 0.46 nJ/inference. The LUTN can be viewed as a highly energy efficient and very fast classification add-on to the ECG preprocessing, in essence reducing the overall power consumption to that of the preprocessing units alone.
    
    \subsection{Comparison with existing ECG classifiers}
    Table \ref{tab:comparison} compares our implementation with other works \linebreak[4] based on energy per inference and power. Similar to several prior ECG arrhythmia classification works \cite{chazal_automatic_2004, mondejar-guerra_heartbeat_2019, yan_energy_2021,mao_ultra-energy-efficient_2022,feng_low-power_2024}, our work utilizes peak positions from the data set itself. Hence, the current utilization, latency and power measurements do not include R-peak detection. Our LUTN IP requires only 0.46 nJ per inference. In comparison, the LGN of Feng et al. \cite{feng_low-power_2024} requires 5.67 nJ per inference. Including preprocessing, our implementation consumes 8250 nJ/inference, while the SNN in \cite{mao_ultra-energy-efficient_2022} only requires 300 nJ/inference. However, we implemented our design on an FPGA instead of an ASIC, meaning that there is additional implementation overhead, and direct comparison is not ideal. Moreover, we employ a much more complex preprocessing, giving rise to a higher accuracy of 94.41\% instead of 93.67\%. The rate-coded SNN$^r$ of \cite{yan_energy_2021} implemented as an ASIC consumes 77 mW without preprocessing, while our FPGA implementation with preprocessing only consumes 2.33 mW. \linebreak[4] Larger models such as an ensemble of SVMs \cite{mondejar-guerra_heartbeat_2019} or a Linear Discriminant (LD) classifier \cite{chazal_automatic_2004} do not report any energy usage. In general, these results highlight that the energy and power value estimations of our LUTN classifier are low compared with existing works. Nevertheless, a direct comparison remains difficult, because of different preprocessing methods and implementation platforms.
    
    \begin{table*}[t]
    \centering
    \caption{Comparison of ECG arrhythmia classifiers in terms of power (mW) and energy/inference (nJ/inf.). The reported energy or power does not always include preprocessing or R-peak detection. $^1$Does not contain any significant preprocessing. $^2$Employs the mixed-patient scheme.}
    \label{tab:comparison}
    \resizebox{\textwidth}{!}{%
    \begin{tabular}{llllllll} 
    \toprule
    \textbf{Classifier} & \textbf{Ref.} & \textbf{Preproc.} & \textbf{R-peak Detec-} & \textbf{Energy/inf.} & \textbf{Power} & \textbf{Process} & \textbf{Platform} \\
     & & \textbf{Included} & \textbf{tion Included} & \textbf{(nJ/inf.)} & \textbf{(mW)} & & \\
    \midrule
    LD classifier & \cite{chazal_automatic_2004} & \textbf{Yes} & No & -- & -- & -- & No implementation \\
    SVM Ensemble & \cite{mondejar-guerra_heartbeat_2019} & \textbf{Yes} & No & -- & -- & -- & No implementation \\
    SNN$^r$ & \cite{yan_energy_2021} & No$^1$ & No & Not reported & 77 & 28 nm & Shenjing \cite{wang_shenjing_2019} ASIC simulator\\
    SNN & \cite{mao_ultra-energy-efficient_2022} & \textbf{Yes} & Not reported & 300 & Not reported & 28 nm CMOS & Fabricated ASIC \\
    LGN & \cite{feng_low-power_2024} & No$^{1,2}$ & No & 5.67 & Not reported &  55 nm CMOS ULP & Synopsys Design Compiler (ASIC)\\
    LUTN & Proposed & No & No & 0.46 & 0.092 & 28 nm HKMG & FPGA implementation\\
    LUTN \& preproc. & Proposed & \textbf{Yes }& No & 8250 & 2.33 & 28 nm HKMG & FPGA implementation\\
    \bottomrule
    \end{tabular}}
    \end{table*}

\section{Conclusion}
We investigated the use of logic-based models, i.e. Deep Differentiable Logic Gate Networks (LGNs) and Lookup Table Networks (LUTNs), as lightweight classifiers for inter-patient ECG arrhythmia classification. A novel LUTN training method was introduced that uses the Boolean equation of a multiplexer, along with preprocessing of the MIT-BIH data set tailored to LGNs and LUTNs. The effect of rate coding, LUT fan-in and network depth on the classification performance was investigated.\\

Our study showed that these logic-based models achieve an accuracy of 94.41\% and a $j\kappa$ index of 0.683, matching CNN-, SVM- and SNN-based  state-of-the-art methods. Yet, our models required only a few thousand FLOPs, which is an estimated three to six orders of magnitude less than state-of-the-art approaches. In addition, our results demonstrated that rate coding LGNs improves the $j\kappa$ index due to an enhanced classification performance of supraventricular ectopic beats. We also demonstrated that 6-LUTNs with 12000 connections per layer obtained a similar performance as 2-LUTNs/LGNs with 16000 connections per layer, enabling a more compact hardware representation. Obtaining an efficient hardware implementation with LGNs and LUTNs is straightforward, since the trained network is directly converted to Verilog, showcasing a one-to-one mapping of 6-LUTNs to FPGA. We implemented our network on FPGA as a verification of our methods. In reality, an ASIC implementation will deliver the lowest achievable energy consumption.\\

Validation of the proposed methods was performed by implementing both the preprocessing and logic-based model on a Xilinx Zynq-7000 FPGA, highlighting the practical feasibility of our approach. The ECG Preprocessing IP occupied \linebreak[4] 17064~LUTs, had a latency of 3540 clock cycles and a dynamic energy per inference of 8.25 µJ. In contrast, the LUTN Classifier IP only needed 3223 LUTs, had a latency of 5 clock cycles and an energy per inference of 0.46 nJ. This confirms that LGNs/LUTNs are promising candidates as lightweight arrhythmia classifiers. The total inference time of both IPs combined was 3.5 ms at a clock speed of 1 MHz, which is ample time to classify heartbeats in streaming mode.\\

Overall, the results show that logic-based classifiers obtain competitive inter-patient ECG classification performance with low computational complexity and a straightforward hardware implementation. This combination suggests that LGNs and \linebreak[4] LUTNs are promising candidates for future resource-con-\linebreak[4]strained biomedical monitoring systems, including implantable or wearable devices, for which current methods are too computationally demanding.\\

A list of possible research directions is given in Table \ref{tab:future_research}. One possible research direction is to adapt the current algorithm to increase the classification performance on the fusion (F) beats on the MIT-BIH data set by employing synthetic minority oversampling (SMOTE) \cite{chawla_smote_2002} and utilizing a focal loss function \cite{lin_focal_2018}. As shown in Table \ref{tab:perclass_metrics}, the supraventricular ectopic (S) beat classification performance is oftentimes lower than state-of-the-art methods, thus low-complexity preprocessing should be added directed at increasing this class-specific performance, which is a second research direction. A third research direction could aim at improving training stability and decreasing training time. Early results have demonstrated that clamping weight values between zero and one at all times increases training stability of deep LUTNs consisting out of high-fan-in LUTs. Alternatively, the trainable LUT entries could be bounded by applying a sigmoid function, limiting their values to $[0,1]$. Sigmoid annealing could be an alternative binarization mechanism to binarize all layers of the network in one go. In contrast, the current algorithm binarizes the network in a layer-wise fashion over the duration of the training. A fourth research direction is the application of convolutional LGNs \cite{petersen_convolutional_2024} and LUTNs to ECG arrhythmia classification. These logic-based convolutional networks could extract non-handcrafted morphological features in an equally hardware-efficient manner, again possibly leading to increased performance. Adding a light-\linebreak[4]weight peak detection algorithm \cite{pan_realtime_1985} is another research direction. This is currently the main component that is missing to obtain an end-to-end arrhythmia classifier. Finally, the LGNs transform input features into class predictions using binary transformations consisting of simple 2-input Boolean operations. Extracting logic rules for specific examples or classes could give insight into \textit{why} the LGN decides to assign a class label to a specific example. The resulting set of Boolean rules could then be compared to clinical protocols, potentially providing insights into clinical decision-making.

\begin{table*}[]
    \centering
    \caption{Possible future research directions and their main motivations.}
    \begin{tabular}{p{0.28\textwidth}|p{0.65\textwidth}}
        \toprule
         Future Research Direction & Motivation\\
         \midrule
         Advanced lightweight preprocessing & The current pipeline does not classify the supraventricular ectopic (S) beats well compared to the state of the art. Therefore, the current preprocessing should be adapted for enhanced S class detection, while keeping the computational complexity simple. \\
         \\
         Advanced class imbalance mitigation & The fusion (F) class is heavily underrepresented in the MIT-BIH data set. To mitigate this underrepresentation, synthetic minority oversampling (SMOTE) \cite{chawla_smote_2002} could be employed to create additional F class examples. In addition, focal loss \cite{lin_focal_2018} could be utilized for better classification performance of examples which are hard to classify.\\
         \\
         Limiting weight values and binarization method & Early results show that clamping LUT entries to $[0,1]$ during the full training phase, enhances final performance of the deep 6-LUTNs on the MIT-BIH data set. Yet, the weight values could be bounded by applying a sigmoid function. Instead of the current layer-wise binarization method that requires more training epochs for deeper networks, sigmoid annealing the full network could be employed to lower training times.\\
         \\
         Convolutional architectures & Instead of utilizing handcrafted morphological features, convolutional LGNs \cite{petersen_convolutional_2024} or convolutional LUTNs could be utilized to extract these automatically. This could lead to a higher performance compared to feedforward architecture in a hardware-efficient manner. \\
         \\
         Peak detection an end-to-end implementation & The main missing component in our setup to obtain and end-to-end ECG arrhythmia classification pipeline, is a lightweight peak detection algorithm such as the Pan-Tompkins algorithm \cite{pan_realtime_1985}. \\
         \\
         Interpretation and explainability & LGNs \cite{yueLearningInterpretableDifferentiable2024} consist of logic gates, and hence for each example or class, one could extract logic rules from input features up until prediction. These rules could be compared with clinical knowledge, providing decision insights for LGNs and clinicians. \\
         \bottomrule
    \end{tabular}
    \label{tab:future_research}
\end{table*}

\section*{CRediT authorship contribution statement}
\textbf{Wout Mommen}: Conceptualization, Data curation, Formal analysis, Investigation, Methodology, Software, Visualization, Writing -- original draft.
\textbf{Lars Keuninckx}: Conceptualization, Data curation, Supervision, Validation, Writing -- review \& editing.
\textbf{Siddharth Patil}: Software, Validation, Visualization.
\textbf{Paul Detterer}: Software, Validation.
\textbf{Achiel Colpaert}: Conceptualization, Supervision, Writing -- review \& editing.
\textbf{Piet Wambacq}: Supervision, Funding acquisition.

\section*{Declaration of competing interest}
The authors declare that they have no known competing financial interests or personal relationships that could have appeared to influence the work reported in this paper.

\section*{Ethics statement}
This study utilized the publicly available MIT-BIH Arrhythmia Database, that has been extensively used in ECG arrhythmia classification benchmarks. The study only involved secondary analysis of anonymized, pre-existing
data and no additional ethical approval or informed consent was required.

\section*{Acknowledgements}
This research received funding from the Flemish Government under
the “Onderzoeksprogramma Artificiële Intelligentie (AI) Vlaanderen” program.

\newpage
\appendix
\section{Validation of proposed methods} \label{sec:validation}
\subsection{Rate coding validation} \label{sec:validation_ratecoding}
\begin{figure}[h!]
     \centering
     \begin{subfigure}[b]{0.4\textwidth}
         \centering
         \includegraphics[width=\linewidth]{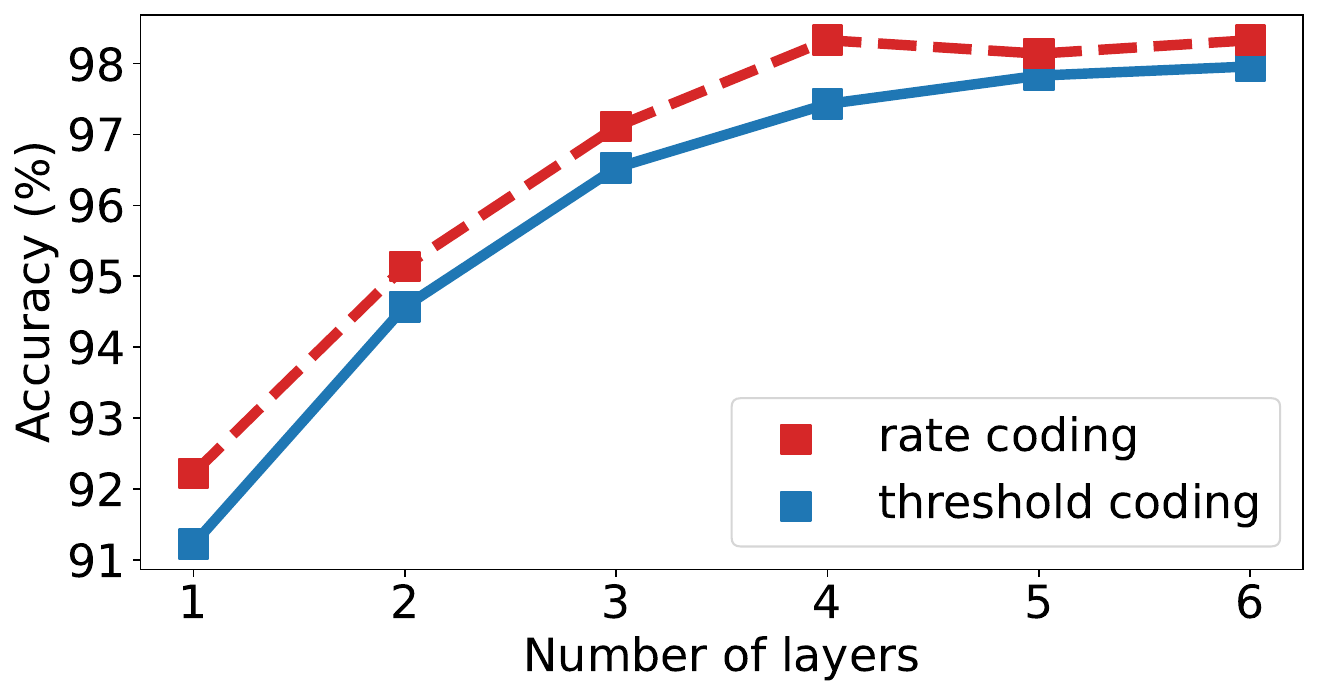}
         \caption{}
         \label{fig:rate_MNIST}
     \end{subfigure}
     \hfill
     \begin{subfigure}[b]{0.4\textwidth}
         \centering
            \includegraphics[width=\linewidth]{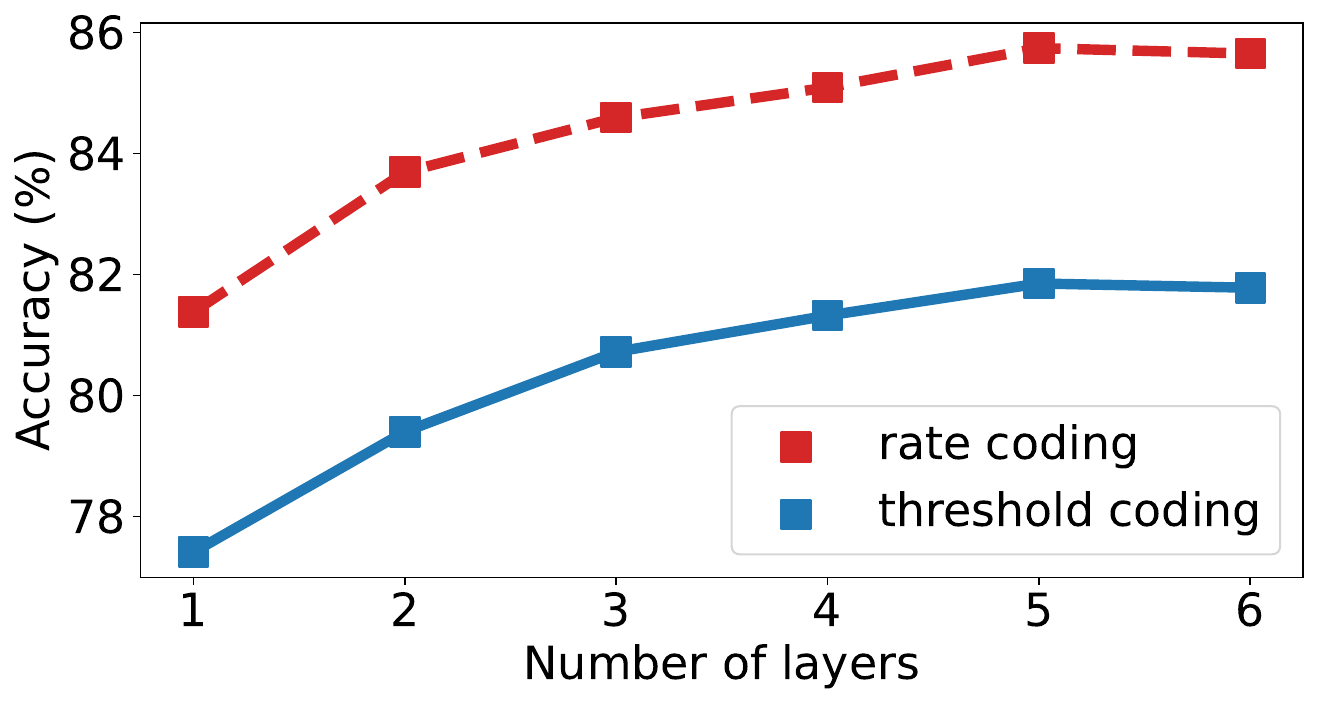}
         \caption{}
         \label{fig:rate_FashionMNIST}
     \end{subfigure}
        \caption{Comparison between single threshold coding and rate coding on (a) the MNIST Handwritten Digits data set and (b) the Fashion-MNIST data set, for LGNs using 8000 gates per layer.}
        \label{fig:rate}
\end{figure}

To evaluate rate coding in LGNs, the networks were tested on the MNIST Handwritten Digits and Fashion-MNIST data sets for different network depths, with 8000 gates per layer. These findings are displayed in Fig. \ref{fig:rate_MNIST} (MNIST Handwritten Digits) and \ref{fig:rate_FashionMNIST} (Fashion-MNIST), and are compared to LGNs with binary input values. The binarized input values are obtained by thresholding the pixel values with a threshold equal to 0.5. Based on these results, all networks with rate-coded inputs outperform the networks with binary input values on both data sets. The results of employing rate coding for the MIT-BIH data set are presented in Section \ref{sec:MIT_BIH}.

\subsection{LUTN training validation} \label{sec:validation_LUTN}
To ensure that the LUTN training method works effectively, it is first tested on the MNIST Handwritten Digits and Fashion-MNIST data sets. Table \ref{tab:LUTN_MNIST} presents the accuracy of an LGN (8000 gates/layer), a 2-LUTN (8000 LUTs/layer), a 4-LUTN (3000 LUTs/layer), and a 6-LUTN (2000 LUTs gates/layer). The 4-LUTNs and 6-LUTNs consist of 12000 connections per layer, while the LGNs and 2-LUTNs contain 16000 connections per layer. All models were also tested with rate-coded input features, indicated by the superscript $r$ in the model name. The 6-LUTNs outperform all other models with binary input features, given a constant number of layers. One exception might be the 2-layer 4-LUTN trained on the Fashion-MNIST data set. This model achieves an accuracy of 80.46\%, which is comparable to the accuracy of the 6-LUTN obtaining 80.00\%. This statement is also true for the networks with rate-coded inputs: The 6-LUTNs$^r$ outperform all other models, given a constant number of layers. Moreover, networks with rate-coded inputs always outperform networks with binary input values, especially on the Fashion-MNIST data set. The reason is that gray-scale values are more important for the Fashion-MNIST data set compared to the MNIST Handwritten Digits data set, where most pixels mostly fall into just two categories, namely black and white. Note that even though the LGNs/LGNs$^r$ contain more connections and number of ``neurons'' compared to networks with 4-input and 6-input LUTs, the 4-LUTNs/4-LUTN$^r$ and 6-LUTNs/6-LUTNs$^r$ always outperform the LGNs/LGNs$^r$. We expect that this is due to the large number of functions an $N$-LUT can represent. As shown in Table \ref{tab:LUT_func}, a 6-LUT can represent $1.845\cdot10^{19}$ functions, while a 2-LUT can only represent 16 different binary functions.

\begin{table}[h!]
    \centering
    \caption{Accuracies (\%) of the LGNs \cite{petersen_deep_2022} and LUTNs on the MNIST Handwritten Digits and Fashion-MNIST data sets.}
    \label{tab:LUTN_MNIST}
    \begin{tabular}{c|cccc}
    \toprule
     &  \multicolumn{2}{c}{MNIST} & \multicolumn{2}{c}{Fashion-MNIST}\\
     \midrule
    Model & 1 layer & 2 layers & 1 layer & 2 layers \\ \hline
    LGN & 91.22 & 94.57 & 77.41 & 79.40 \\
    2-LUTN & 91.74 & 94.22 & 77.36 & 79.20 \\
    4-LUTN & 92.66 & 96.73 & 78.19 & \textbf{80.46} \\
    6-LUTN & \textbf{93.88} & \textbf{97.29} & \textbf{79.61} & 80.00 \\
    \midrule
    LGN$^r$ & 91.22 & 95.14 & 81.38 & 83.69 \\
    2-LUTN$^r$ & 92.26 & 94.85 & 81.83 & 83.75 \\
    4-LUTN$^r$ & 94.30 & 97.85 & 83.97 & 86.17 \\
    6-LUTN$^r$ & \textbf{95.34} & \textbf{98.41} & \textbf{85.09} & \textbf{87.48} \\
    \bottomrule
    \end{tabular}
\end{table}

\section{Additional MIT-BIH results}
\subsection{Optimal softmax temperatures} \label{sec:optimal_temp}
The performance of LGNs and LUTNs depends on the chosen hyperparameter values. One such hyperparameter is the output softmax temperature, that converts the population counts to class probabilities. The optimal output softmax temperature is defined as the temperature value for which the network attains the highest accuracy on the data set, while keeping all other hyperparameter values fixed. Table \ref{tab:Mixed_patients_temp} displays the optimal softmax temperature values for the LGNs on the MIT-BIH data set when patients are mixed in train and test set. Table \ref{tab:temp_inter_patient} presents the optimal softmax temperature values of the LGN and LUTNs on the MIT-BIH data set employing the inter-patient paradigm. In both cases, this temperature was found utilizing a grid search in the range of 1 to 10, for all integer values. For values above 10, a step size of 5 was employed. As mentioned in \cite{petersen_deep_2022}, networks with wide layers obtain a higher final accuracy if a larger temperature is chosen. This trend is also observed in Table \ref{tab:temp_inter_patient} for the LUTNs. The optimal temperature value for 6-LUTNs (2000 LUTs/layer) is 25, for 4-LUTNs (3000 LUTs/layer) is 30, and for 2-LUTNs (8000 LUTs/layer) is 50. In addition, the optimal temperature value of 6-LUTNs$^r$ (2000 LUTs/layer) is 15, for 4-LUTNs$^r$ (3000 LUTs/layer) is 40, and for 2-LUTNs$^r$ (8000 LUTs/layer) is 50. The superscript $r$ in the model name indicates rate-coded network inputs.
\begin{table}[H]
    \centering
    \caption{Ideal softmax temperatures for the networks that use mixed patients for the MIT-BIH data set. Our models use the preprocessing as explained in Section \ref{sec:feature_extraction}, while Feng et al. \cite{feng_low-power_2024} utilizes dynamic thresholds. The superscript $r$ indicates rate-coded inputs. All LGNs consisted out of 8000 gates per layer.}
    \label{tab:Mixed_patients_temp}
    \begin{tabular}{lccc}
    \toprule
    Model & Ref. & No. classes & Temp.\\
    \midrule
     LGNs & Proposed & 4 & 30\\
     LGNs$^r$ & Proposed & 4 & 20\\ 
     LGNs & \cite{feng_low-power_2024} & 4 & 8\\   
     LGNs & \cite{feng_low-power_2024} & 5 & 5\\   
    \bottomrule
    \end{tabular}
\end{table}

\begin{table}[h!]
    \centering
    \caption{Ideal softmax temperatures for the networks that use the inter-patient paradigm. Our models use the preprocessing as explained in Section \ref{sec:feature_extraction}, while Feng et al. \cite{feng_low-power_2024} utilizes dynamic thresholds. The superscript $r$ indicates rate-coded inputs. The LGNs and 2-LUTNs had 8000 gates/LUTs per layer, while the 4-LUTNs and 6-LUTNs consisted of 3000 and 2000 LUTs per layer respectively.}
    \label{tab:temp_inter_patient}
    \begin{tabular}{lcc}
    \toprule
    Model & Ref. & Temp.\\
    \midrule
     LGNs & \cite{feng_low-power_2024} & 30\\   
     LGNs & Proposed & 35\\
     LGNs$^r$ & Proposed & 10\\ 
     \midrule
     2-LUTNs & Proposed & 50\\ 
     4-LUTNs & Proposed & 30\\ 
     6-LUTNs & Proposed & 25\\
     \midrule
     2-LUTNs$^r$ & Proposed & 50\\
     4-LUTNs$^r$ & Proposed & 40\\
     6-LUTNs$^r$ & Proposed & 15\\ 
    \bottomrule
    \end{tabular}
\end{table}

\subsection{Detailed performance results} \label{sec:detailed_results}
Table \ref{tab:Acc_inter_patient} and Table \ref{tab:jk_inter_patient} present the accuracy and $j\kappa$ index of the LGNs and LUTNs employing binary input features and rate coding. The models are trained on the MIT-BIH test set utilizing the inter-patient paradigm.\\

\textbf{LGNs vs 2-LUTNs} Table \ref{tab:Acc_inter_patient} and Table \ref{tab:jk_inter_patient} show that the 2-LUTNs outperform LGNs based on the accuracy and $j\kappa$ index for all depths. Hence for binary input features, the 2-LUTNs are preferred, due to a better performance. The primitives of LGNs and 2-LUTs can learn exactly the same functions, but an LGN primitive needs 16 trainable parameters, while a 2-LUT primitive only needs four, reducing the number of training parameters by a factor of four. Although, the 2-LUTNs perform better than LGNs with binary input values, the opposite is true when the input data is rate-coded.\\

\textbf{Rate coding and LGNs} Based on the accuracy and $j\kappa$ index results in Table \ref{tab:Acc_inter_patient} and Table \ref{tab:jk_inter_patient} respectively, we observe that the accuracy is higher and the $j\kappa$ index is lower for the LGNs compared to the rate-coded LGNs$^r$. For the LGNs$^r$, the number of correctly classified supraventricular ectopic (S) beats is higher, while the number of correctly classified normal (N) beats is lower compared to the LGNs. Since the performance is better on the S class, the $F_1$ score of the LGNs$^r$ in Table \ref{tab:perclass_metrics} is higher compared to the one of the LGNs. Thus, the values of the $j$ index and $j\kappa$ index of LGNs$^r$ are also higher than the ones of LGNs. Still, for the LGNs$^r$, the decrease in performance on the normal class is larger in magnitude than the increase in performance on the S class compared to the LGNs. This is the reason why the accuracy of the LGNs$^r$ are lower than the accuracy of the LGNs. To conclude: Rate coding our input features enhances the performance on the S class, while diminishing the performance for normal heartbeats. A detailed investigation into the classification performance of the supraventricular ectopic beats is necessary to pinpoint why this is the case. This corresponds to the first entry in Table \ref{tab:future_research}, which lists possible future research directions.\\

\textbf{Depth of LGNs} The accuracy of the LGNs rises with the number of layers, as seen in Table \ref{tab:Acc_inter_patient}. Based on the obtained confusion matrices, we conclude that the normal class and ventricular ectopic (V) class are predicted correctly more often. The opposite is true for networks with rate-coded inputs: The accuracy decreases with depth for the LGNs$^r$. For example, a 2-layer network predicts the normal class less well than a single layer network. The decrease in accuracy from 2-layer networks to 4-layer networks is due to a worse performance on the S class, which also explains the downwards trend of the $j\kappa$ index. A careful analysis of the S class performance is required to explain this downwards trend. This is the first mentioned future research direction in Table \ref{tab:future_research}.\\

\textbf{LUT fan-in} Table \ref{tab:Acc_inter_patient} and Table \ref{tab:jk_inter_patient} show that the accuracy and $j\kappa$ index are lower for deep 6-LUTNs compared to deep 4-LUTNs and deep 2-LUTNs. In detail, the 4-layer 6-LUTN achieves an accuracy of 88.67\% and a $j\kappa$ index of 0.000. Nevertheless, the training method does not force the weight values to lie within $[0,1]$ before binarization. Early results in Table \ref{tab:clamping} show that if these weights are clamped throughout the full training phase, the accuracy and $j\kappa$ index remain stable. Namely, a 4-layer 6-LUTN now obtains an accuracy of 94.32\% and a $j\kappa$ index of 0.659.\\

\textbf{Rate coding and LUTNs} Utilizing rate coding with the LUTNs decreases both accuracy and $j\kappa$ index for all considered LUT fan-in values compared to using binary input values. Utilizing the above clamping throughout the full training phase, of which the results are shown in Table \ref{tab:clamping}, did not improve the results shown in Table \ref{tab:Acc_inter_patient} and Table \ref{tab:jk_inter_patient}. Yet, the $j\kappa$ index of the 4-layer 6-LUTN$^r$ is now equal to 0.552 instead of 0.000. To assess if the cause of the lower accuracy and $jk$ index was due to the LUTN$^r$ model or due to the input features themselves, a multilayer perceptron (MLP) was trained on the binarized and one on continuous-valued inputs. Each MLP consisted of two hidden layers of 256 and 128 neurons respectively, with ReLU activation functions. Both the MLPs with binary inputs and with continuous-valued inputs were trained 30 times, each with a different seed. On average, the MLP with binarized input values obtains an accuracy of $94.11\pm0.18\%$ and a $j\kappa$ index of $0.669\pm0.012$. The MLP with continuous-valued inputs obtains an accuracy of $93.09\pm 0.32\%$ and a $j\kappa$ index of $0.608\pm 0.050$. Clearly, the MLP with binary input values performs better than the MLP employing continuous-valued inputs. It is plausible that binarizing the input features acts as a regularization mechanism that suppresses patient-specific variability. Yet, the reason why the LGNs with rate-coded inputs obtain a higher $j\kappa$ index compared to the LGN with binarized inputs is unknown at the time of writing.\\

\textbf{Depth of LUTNs} The accuracy increases with depth for 2-LUTNs, stays more or less constant for 4-LUTNs, and decreases for 6-LUTNs. When clamping is utilized, the accuracy of the 6-LUTNs does not decrease with depth anymore, as is shown in Table \ref{tab:clamping}. The $j\kappa$ index also shows a decrease for 3-layer and 4-layer 6-LUTNs, but this has been resolved by clamping weight values to the interval $[0,1]$.

\begin{table*}[!h]
    \centering
    \caption{Accuracies (\%) of our networks (LGNs, 2-LUTNs, 4-LUTNs, and 6-LUTNs) that use the inter-patient paradigm for different number of layers. The LGNs/2-LUTNs have 8000 gates/LUTs per layer, the 4-LUTNs have 3000 LUTs/layer, and the 6-LUTNs have 2000 LUTs/layer. The superscript $r$ indicates the use of rate coding as described in this work. This is compared to the known state-of-the-art LGN results on this data set. The LUT entries are not clamped to $[0,1]$ during training.}
    \label{tab:Acc_inter_patient}
    \begin{tabular}{cccccccccc}
    \toprule
    No. Layers & LGNs & LGNs$^r$ & 2-LUTNs & 2-LUTNs$^r$ & 4-LUTNs & 4-LUTNs$^r$ & 6-LUTNs & 6-LUTNs$^r$ & LGNs \cite{feng_low-power_2024}\\
    \midrule
    1 & 93.79 & 93.63 & 93.99 & 93.17 & 94.00 & 92.91 & 94.04 & 92.66 & 90.86 \\
    2 & 94.06 & 93.48 & 94.18 & 93.34 & 94.26 & 92.85 & 94.24 & 92.79 & 90.30 \\
    3 & 94.19 & 93.50 & 94.23 & 93.41 & 94.21 & 93.14 & 94.20 & 93.17 & 89.73 \\
    4 & 94.28 & 93.26 & \textbf{94.41} & 93.54 & 94.23 & 93.20 & 88.67 & 92.14 & 91.13 \\
    \bottomrule
    \end{tabular}
\end{table*}

\begin{table*}[!h]
    \centering
    \caption{$j\kappa$ indexes of our networks (LGNs, 2-LUTNs, 4-LUTNs and 6-LUTNs) that use the inter-patient paradigm for different number of layers. The LGNs/2-LUTNs have 8000 gates/LUTs per layer, the 4-LUTNs have 3000 LUTs/layer and the 6-LUTNs have 2000 LUTs/layer. The superscript $r$ indicates the use of rate coding as described in this work. This is compared to the known state-of-the-art LGN results on this data set. The LUT entries are not clamped to $[0,1]$ during training.}
    \label{tab:jk_inter_patient}
    \begin{tabular}{cccccccccc}
    \toprule
    No. layers & LGNs & LGNs$^r$ & 2-LUTNs & 2-LUTNs$^r$ & 4-LUTNs & 4-LUTNs$^r$ & 6-LUTNs & 6-LUTNs$^r$ & LGNs \cite{feng_low-power_2024}\\
    \midrule
    1 & 0.645 & 0.633 & 0.659 & 0.595 & 0.646 & 0.549 & 0.651 & 0.572 & 0.407 \\
    2 & 0.569 & \textbf{0.683} & 0.669 & 0.618 & 0.638 & 0.614 & 0.643 & 0.571 & 0.434 \\
    3 & 0.650 & 0.660 & 0.654 & 0.585 & 0.624 & 0.557 & 0.583 & 0.561 & 0.412 \\
    4 & 0.559 & 0.649 & 0.669 &  0.610 & 0.635 & 0.612 & 0.000 & 0.000 & 0.409 \\
    \bottomrule
    \end{tabular}
\end{table*}

\begin{table}[t]
\centering 
\caption{Accuracy (\%) and $j\kappa$ index on the inter-patient MIT-BIH data set for 6-LUTNs and rate-coded 6-LUTNs (6-LUTNs$^r$) using 2000 6-LUTs per layer. These results applied clamping of the LUT entries to $[0,1]$ during training.}
\begin{tabular}{c|cc|cc} 
\toprule 
& \multicolumn{2}{c|}{6-LUTNs} & \multicolumn{2}{c}{6-LUTNs$^r$} \\
\midrule
No. Layers & Accuracy & $j\kappa$ & Accuracy & $j\kappa$ \\
\midrule 
1 & 94.07 & 0.649 & 92.59 & 0.565 \\ 
2 & 94.30 & 0.655 & 92.67 & 0.562 \\ 
3 & 94.36 & 0.647 & 93.09 & 0.537 \\ 
4 & 94.32 & 0.645 & 92.80 & 0.552 \\ 
\bottomrule \end{tabular} 
\label{tab:clamping} 
\end{table}

\subsection{Convergence of rate coding} \label{sec:conv_rate_coding}
\begin{figure}
    \centering
    \includegraphics[width=\linewidth]{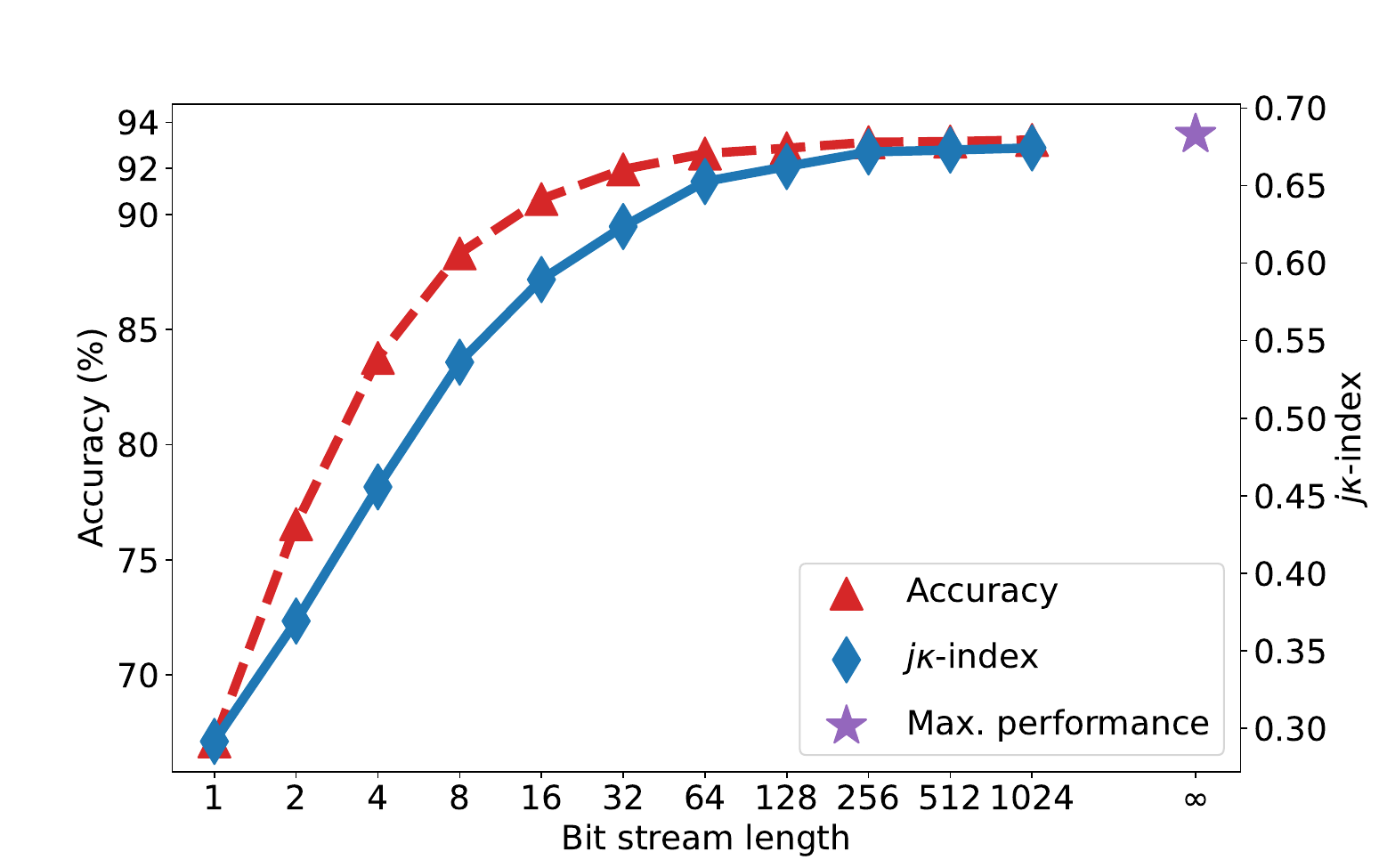}
    \caption{The accuracy and $j\kappa$ index of a 2-layer rate-coded LGN for a varying number of bit stream lengths. As expected, the longer the bit stream length, the closer our classifier is to attaining the maximum possible performance.}
    \label{fig:acc_jk_rate}
\end{figure}

Fig. \ref{fig:acc_jk_rate} displays the accuracy and $j\kappa$ index for a 2-layer LGN$^r$ utilizing different bit stream lengths. The longer the bit stream length, the higher the accuracy and $j\kappa$ index. The accuracy and $j\kappa$ index values converge to the LGN$^r$ values in Table~\ref{tab:inter_patient_summary}, corresponding with the confusion matrix in Table \ref{tab:confusion_matrix}, when a long enough bit stream length is applied. This behaviour is expected because LGNs$^r$ are trained with full-precision values, corresponding to theoretically infinite bit stream lengths.

\clearpage

\bibliographystyle{elsarticle-num}  
\bibliography{references}           
\end{document}